\documentclass[journal]{IEEEtran}
\ifCLASSINFOpdf
\else
\fi
\usepackage[cmex10]{amsmath}
\usepackage{array}

\usepackage{mdwmath}
\usepackage{mdwtab}

\usepackage{eqparbox}
\usepackage{fixltx2e}
	\usepackage{url}

 \usepackage{hyperref}
\usepackage{float}
\usepackage{color}
\usepackage[noadjust]{cite}
\usepackage{graphicx}
\usepackage{tikz,pgfplots}
\usepackage{subfigure}
\usepackage{graphicx}
\usepackage{epstopdf}
\usepackage{amssymb,mathrsfs}
\usepackage{booktabs}
\usepackage{multirow}
\usepackage{multicol}
\usepackage{color}
\usepackage[ruled,vlined]{algorithm2e}

\begin{document}
%
\title{Feature Extraction for Hyperspectral Imagery: \\The Evolution from Shallow to Deep (Overview and Toolbox)}
%
%
%

\author{Behnood~Rasti,~\IEEEmembership{Senior Member,~IEEE,} Danfeng~Hong,~\IEEEmembership{Member,~IEEE,}
Renlong~Hang,~\IEEEmembership{Member,~IEEE,}
Pedram~Ghamisi,~\IEEEmembership{Senior Member,~IEEE,}
Xudong~Kang,~\IEEEmembership{Senior Member,~IEEE,}
Jocelyn~Chanussot,~\IEEEmembership{Fellow,~IEEE, and}
Jon Atli Benediktsson,~\IEEEmembership{Fellow,~IEEE}

\thanks{This work is partially supported by Alexander von Humboldt (AvH) research grant.}
\thanks{B. Rasti is with the Machine Learning Group, Exploration Division, Helmholtz Institute Freiberg for Resource Technology, Helmholtz-Zentrum Dresden-Rossendorf, 09599 Freiberg, Germany (e-mail: behnood.rasti@gmail.com).}
\thanks{D. Hong is with Univ. Grenoble Alpes, CNRS, Grenoble INP, GIPSA-lab, 38000 Grenoble, France. (e-mail: hongdanfeng1989@gmail.com)}
\thanks{R. Hang is with the Jiangsu Key Laboratory of Big
Data Analysis Technology, School of Automation, Nanjing University
of Information Science and Technology, Nanjing 210044, China (e-mail:renlong\_hang@163.com)}
\thanks{P. Ghamisi is with the Machine Learning Group, Exploration Division, Helmholtz Institute Freiberg for Resource Technology, Helmholtz-Zentrum Dresden-Rossendorf, 09599 Freiberg, Germany. (e-mail: p.ghamisi@gmail.com)}
\thanks{X. Kang is with the College of Electrical and Information Engineering, Hunan University, Changsha 410082, China, and also with the Key Laboratory of Visual Perception and Artificial Intelligence of Hunan Province,
Changsha 410082, China (e-mail: xudong\_kang@163.com).}
\thanks{J. Chanussot is with the Univ. Grenoble Alpes, Inria, CNRS, Grenoble INP, LJK, F-38000 Grenoble, France, also with the Faculty of Electrical and Computer Engineering, University of Iceland, Reykjavik 101, Iceland. (e-mail:  jocelyn@hi.is)}
\thanks{J. A. Benediktsson is with the Faculty of Electrical and Computer Engineering, University of Iceland, 107 Reykjavik, Iceland (e-mail:  benedikt@hi.is).}

}
%
%

\markboth{IEEE GRSM DRAFT 2020}%
{Shell \MakeLowercase{\textit{et al.}}: Bare Demo of IEEEtran.cls for Journals}
%



\maketitle

\begin{abstract}

\textcolor{blue}{The final version of the paper can be found in IEEE
Geoscience and Remote Sensing Magazine.} 
Hyperspectral images provide detailed spectral information through hundreds of (narrow) spectral channels (also known as dimensionality or bands) with continuous spectral information that can accurately classify diverse materials of interest. The increased dimensionality of such data makes it possible to significantly improve data information content but provides a challenge to the conventional techniques (the so-called curse of dimensionality) for accurate analysis of hyperspectral images. Feature extraction, as a vibrant field of research in the hyperspectral community, evolved through decades of research to address this issue and extract informative features suitable for data representation and classification. The advances in feature extraction have been inspired by two fields of research, including the popularization of image and signal processing as well as machine (deep) learning, leading to two types of feature extraction approaches named shallow and deep techniques. This article outlines the advances in feature extraction approaches for hyperspectral imagery by providing a technical overview of the state-of-the-art techniques, providing useful entry points for researchers at different levels, including students, researchers, and senior researchers, willing to explore novel investigations on this challenging topic. 
In more detail, this paper provides a bird's eye view over shallow (both supervised and unsupervised) and deep feature extraction approaches specifically dedicated to the topic of hyperspectral feature extraction and its application on hyperspectral image classification. Additionally, this paper compares 15 advanced techniques with an emphasis on their methodological foundations in terms of classification accuracies. Furthermore, to push this vibrant field of research forward an impressive amount of codes and libraries is shared at: \href{https://github.com/BehnoodRasti/HyFTech-Hyperspectral-Shallow-Deep-Feature-Extraction-Toolbox}{\textcolor{blue}{https://github.com/BehnoodRasti/HyFTech-Hyperspectral-Shallow-Deep-Feature-Extraction-Toolbox}}.

\end{abstract}


%
\IEEEpeerreviewmaketitle

\section{Introduction}
\label{Sec:Intro}

\IEEEPARstart{H}{yperspectral} imaging technology provides detailed spectral information by sampling the reflective portion of the electromagnetic spectrum covering a wide range from the visible region (0.4-0.7 $\mu$m) to the short-wave infrared (SWIR) region (almost 2.4 $\mu$m). Hyperspecral sensors can also characterize the emissive properties of objects by acquiring data in the range of the mid-wave and long-wave infrared regions, in hundreds of narrow contiguous spectral channels.

Detailed spectral information provided by hyperspectral sensors present both challenges and opportunities. For instance, hyperspectral images can be used to differentiate between different classes of interest with slightly different spectral characteristics \cite{7882742}.  However, most of the commonly
used methods utilized for the analysis of gray scale, color, or multispectral
images cannot be extended to analyse hyperspectral images due to several reasons, which will be detailed later in Section I.A.

\begin{figure*}
	  \centering
			\includegraphics[width=1\textwidth]{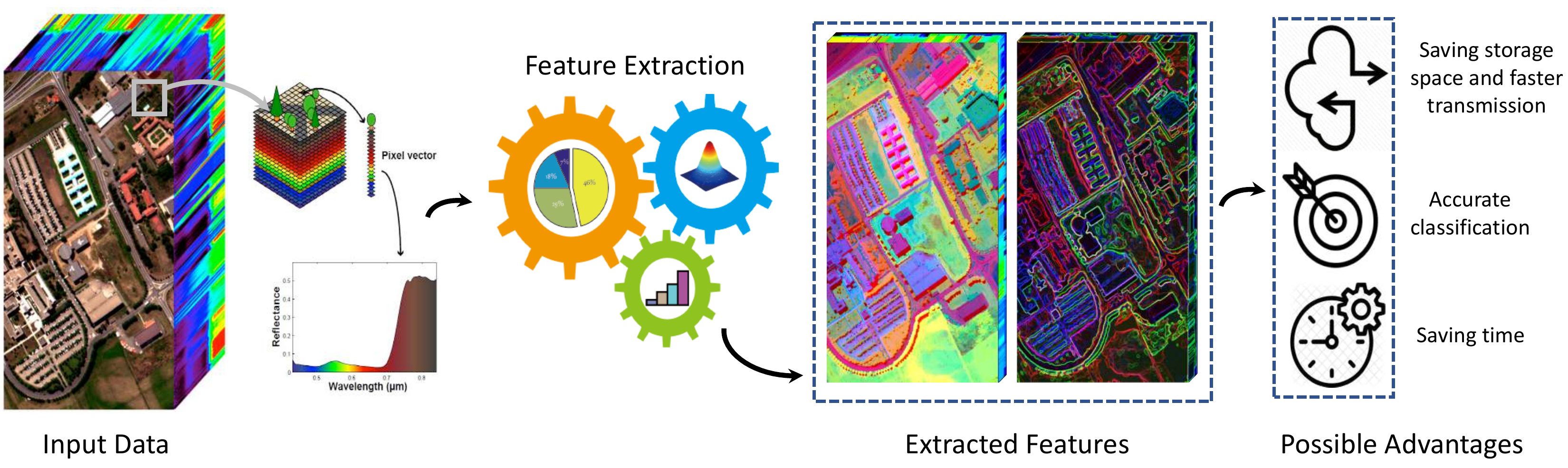}
        \caption{Feature extraction and its advantages for hyperspectral image analysis.}
\label{fig:FE-benefits}
\end{figure*}

The limited availability of training samples (which is a common issue in remote sensing) dramatically impacts the performances of supervised classification approaches due to the high dimensionality of hyperspectral images, which poses a problem for designing robust statistical estimations.
Feature extraction can be used as a solution to address the aforementioned issue, which can be described as finding a set of vectors that represents an observation while reducing the dimensionality by transforming the input data linearly or nonlinearly to another domain and extract informative features in the new domain. The use of feature extraction techniques can be advantageous for a number of reasons which is illustrated in Fig. \ref{fig:FE-benefits} and described in the following subsections.

\subsection{Unique Geometrical and Statistical Properties of High Dimensional Data and the Need for Feature Extraction}

Several studies (e.g., \cite{15,16,PropertyHD}) have demonstrated the unique geometrical, statistical, and asymptotical properties of high-dimensional data compared to RGB and multispectral images. These properties, which have been shown through some experimental and theoretical examples, clearly justify the reasons why most of the analytical approaches that are developed for RGB and multispectral images are not applicable to hyperspectral images \cite{BenediktssonGhamisiBOOK}. Among those experimental examples, we can recall that (1) as dimensionality increases, the volume of a hypercube concentrates in corners, or (2) as dimensionality increases, the volume of a hypersphere concentrates in an outside shell. With respect to these examples, the following conclusions have been drawn:
\begin{itemize}
\item A high-dimensional feature space is almost empty, which indicates that multivariate data in $\mathbb{R}^{p}$ ($p$ represents the number of bands, spectral channels, or dimensions)
can usually be represented in a lower dimensional space (referred to as subspace) without loosing considerable information in terms of class separability \cite{BenediktssonGhamisiBOOK}.
\item Since the high-dimensional feature space is almost empty (i.e., Gaussian distributed data have a tendency to concentrate in the tails while uniformly distributed data have a tendency to be concentrated in the corners), the density estimation of hyperspectral data for both Gaussian and uniform distributions become extremely challenging.
\end{itemize}

Fukunaga \cite{Fukunaga} claimed that there is a relation between the type of the classifier, the required number of training samples, and the number of input dimensions. As reported in \cite{Fukunaga}, the required number of training samples is linearly related to the dimensionality for linear classifiers and to the square of the dimensionality for quadratic classifiers (e.g., the Gaussian maximum likelihood classifier~\cite{Fukunaga}), and for nonparametric classifiers, the number of required samples exponentially increases as the dimensionality increases. Landgrebe showed a ground-breaking fact that too many spectral bands might have negative impacts in terms of expected classification performance \cite{LandgrebeBook}. When dimensionality increases, with a constant and limited number of training samples, a higher amount of statistics must be estimated. Thus, the accuracy of the statistical estimation decreases although higher spectral dimensions increase the separability between the classes. This leads to a decrease in classification accuracies beyond an unknown number of bands. These problems are related to the \textit{curse of dimensionality}, also known as \textit{Hughes phenomenon} \cite{Hughes}. This finding was against the general understanding of hyperspectral data where it was wrongly believed that full dimensionality is always better than subspace in terms of classification accuracies.

The unique characteristics of high-dimensional data, as discussed above, have a pronounced impact on the performances of supervised classifiers \cite{Qian2009bs}, as they demand an adequate number of training samples, which is almost impossible to obtain in practice since the collection of such training samples is either expensive or time demanding. To address this issue, feature extraction-based dimensionality reduction is found to be effective.

\subsection{Storage Systems and Processing Times}

We are now in the era of massive data acquisitions. Statistics demonstrate that the cumulative volume of existing big data has been tremendously increased from 4.4 ZB to 44 ZB from 2013 to 2020\footnote{https://www.emc.com/leadership/digital-universe/2014iview/executive-summary.htm}. The EO community has also faced a similar trend because of the enormous volume and variety of data being generated by EO missions. For example, EnMAP (Environmental Monitoring and Analysis Program), which is a hyperspectral satellite mission, is planning to capture hyperspectral data with a maximum ground coverage of 5000 km $\times$ 30 km per day and the target revisit time of 4 days ($\pm 30^{\circ}$) with 512 Gbit onboard mass memory \cite{EnMAP}.

Feature extraction-based dimensionality reduction helps in data compression, which leads to the reduced storage space, faster transmission time, removing redundant features, reducing the storage space required, and fasten the required time for performing the same computations.

\subsection{An Ever-Growing Relation between Machine learning and Feature Extraction}

\begin{figure}[!t]
	  \centering
		\subfigure{
			\includegraphics[width=0.48\textwidth]{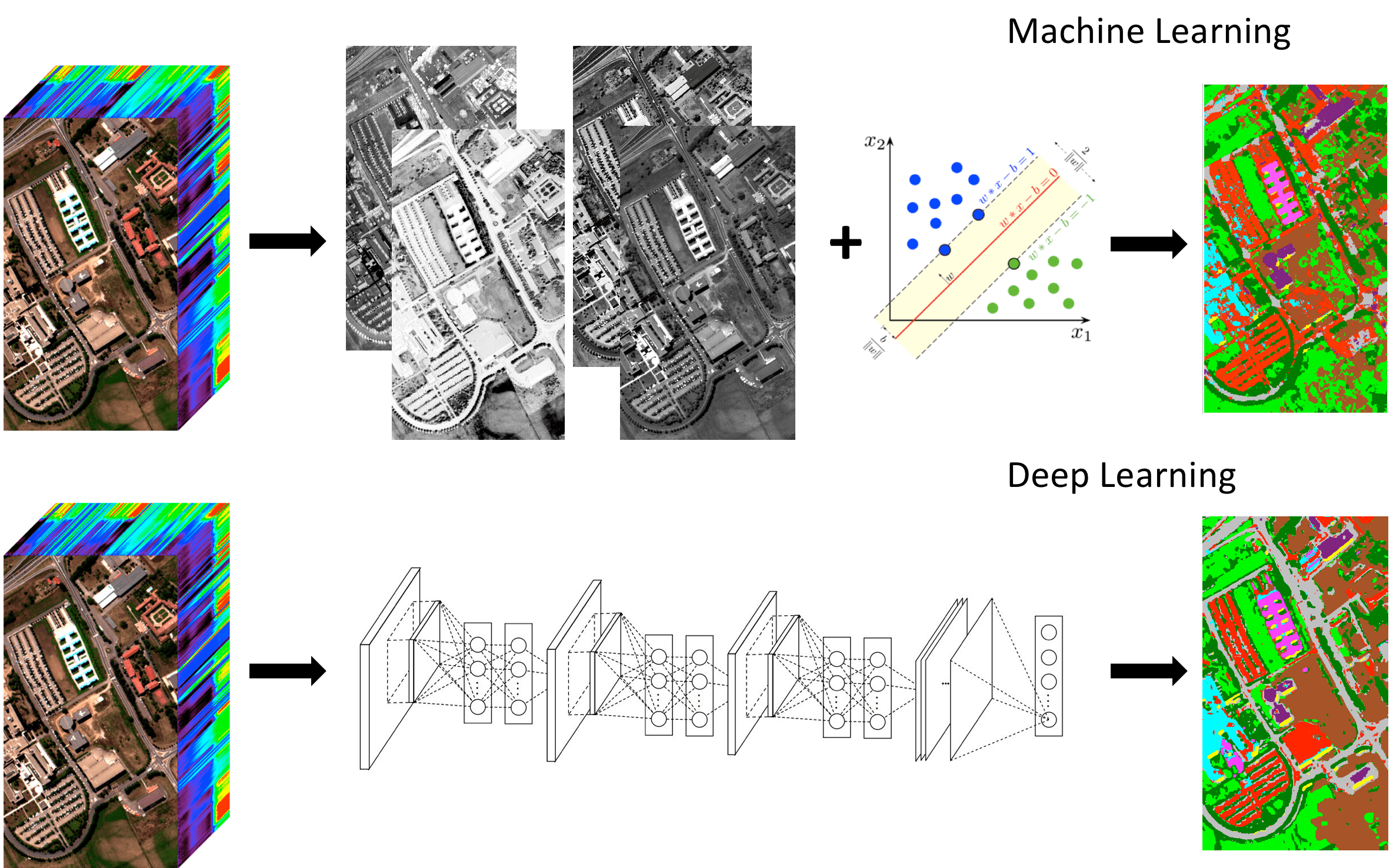}}
        \caption{Feature extraction via machine learning and deep learning.}
\label{fig:machinevsdeep}
\end{figure}

Fig.~\ref{fig:machinevsdeep} illustrates the basic idea of a machine learning approach, which consists of  feature extraction and classification. In machine learning, users are requested to provide some guidelines for the machine (algorithm). These guidelines are usually provided by applying hand-crafted feature extraction approaches to provide informative features for the subsequent classifier. At the very beginning, each image pixel is regarded as a pattern, and its spectrum (i.e., a vector of different values of a pixel in different spectral channels) is considered as the initial set of features. This set of features, which are also known as spectral features, suffers from two important downsides: (1) These features are often redundant and (2) they do not  consider the spatial dependencies of the adjacent pixels.

To address the first issue, a feature reduction (through feature extraction or selection) step can be applied, aiming at reducing the dimensionality of the input data (from $p_{1}$ dimensions in the original data to $p_{2}$ dimensions in a new feature space $p_{2} < p_{1}$). This step can also be known as \textit{spectral feature extraction}, which tries to preserve the key spectral information of the data by reducing the dimensionality and maximizing separability between classes. It is interesting to note that the second issue can also be addressed using feature extraction approaches. Please note that here the feature extraction step, which is also known as \textit{spatial feature extraction}, is not aiming at reducing the dimensionality and instead is trying to model (extract) spatial contextual information suitable for the subsequent classification or object detection step and usually leads to an increase in the number of features. The simultaneous use of spectral and spatial features for hyperspectral data classification were studied in numerous works such as \cite{BenediktssonGhamisiBOOK,8474403,LZ_2018,Fauvel13}.

Deep learning (as shown in Fig.~\ref{fig:machinevsdeep}), which is regarded as a subset of machine learning, tries to automatize the building blocks of the machine learning approaches (i.e., feature extraction and classification) by developing an end-to-end framework, which takes the input, performs automatic feature extraction and classification by considering the unique nature of the input data (instead of those hand-crafted feature extraction designs in machine learning), and outputs classification maps. It turns out that if an adequate amount of training data is supplied, deep learning approaches can outperform any other shallow machine learning approaches in terms of accuracies. Here, a question arises: Due to the fact that the number of available training data is often limited in the remote sensing community, would advanced deep learning-based approaches outperform their shallow alternatives in terms of accuracies? This question will be addressed in this paper.

Based on the above-mentioned descriptions, feature extraction is a key step in both machine learning and deep learning, whose concept has been evolved significantly through time from unsupervised to (semi-)supervised, from spectral or spatial to spectral-and-spatial, from manual to automatic, from hand-crafted to end-to-end, and from shallow to deep.

\subsection{Contributions}

This article provides a detailed and organized overview of hyperspectral feature extraction techniques, categorized into two general sections: shallow feature extraction techniques (further categorized into supervised and unsupervised) and deep feature extraction techniques. Each section provides a critical overview of the state of the art that is mainly rooted in the signal and image processing, statistical inference, and machine (deep) learning fields. Then, a few representative and advanced feature extraction approaches are chosen from each of the above-mentioned categories for further analysis and comparisons (mostly in terms of usefulness for classification). This article will, therefore, contribute to answering the following questions: When it comes to hyperspectral data in Earth observation, are deep learning-based feature extraction approaches better alternatives than their conventional (yet advanced) shallow feature extraction techniques? Which factors should be considered to design robust shallow and deep feature extraction techniques? In addition, to further promote this field of research, this paper is accompanied with a significant amount of codes and libraries for hyperspectral feature extraction, which is made publicly available at \href{https://github.com/BehnoodRasti/HyFTech-Hyperspectral-Shallow-Deep-Feature-Extraction-Toolbox}{\textcolor{blue}{https://github.com/BehnoodRasti/HyFTech-Hyperspectral-Shallow-Deep-Feature-Extraction-Toolbox}}. Finally, several possible future directions are highlighted. To make the contribution of this paper clearer compared to the existing papers in the literature, here we provide a brief discussion. The work of \cite{WeiLi_2018} is  dedicated to the evolution of discriminant analysis-based feature extraction models, which is a specific type of dimensionality reduction approaches. The work of \cite{Fea_mining} reviewed feature extraction and data mining works, which had been published mostly until 2012. Since 2012, however, many deep and shallow feature extraction approaches have been developed, which are critically reviewed and compared against each other in this work. The work of \cite{W_Sun2019} focuses only on feature selection approaches while our proposed paper covers feature extraction techniques, and therefore, they complement each other.

\section{Datasets and Notations}
\label{sec: D and N}

\subsection{Datasets}
\subsubsection{Indian Pines 2010}
\begin{figure} [tbp]\begin{center}
 \includegraphics[width=0.99\linewidth]{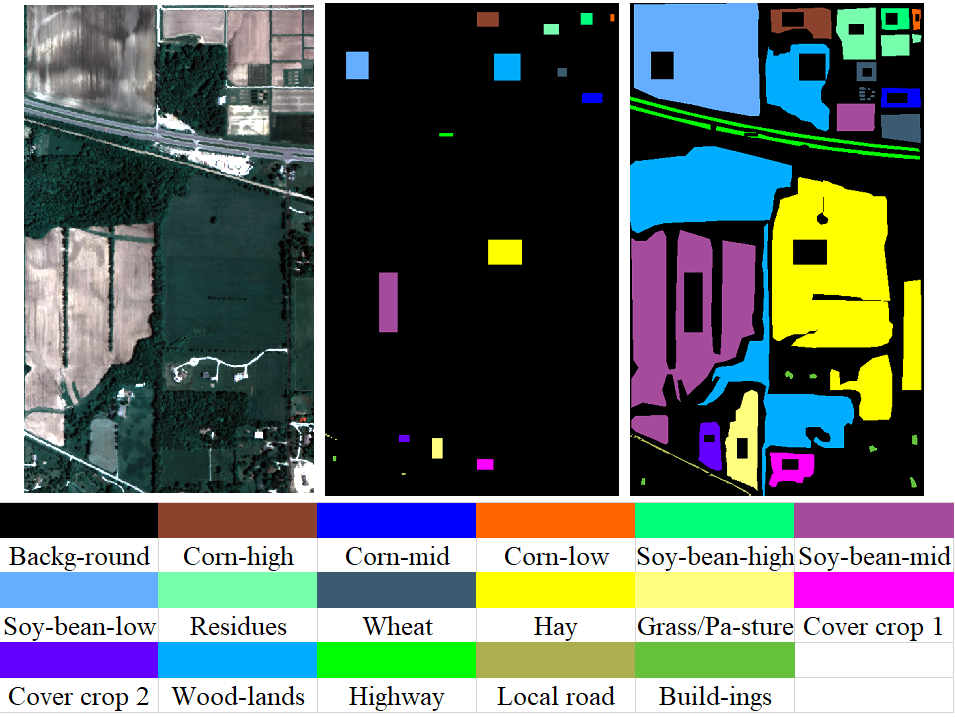}
 \end{center}
 \caption{Indian Pines 2010 datasets, from left to right, RGB composition, Training set, and Test set.}
  \label{fig:Indian10}
 \end{figure}

This dataset (Fig. \ref{fig:Indian10}) is a very high resolution hyperspectral image (VHR HS) acquired by the ProSpecTIR system over near Purdue University, Indiana, between 24-25th of May 2010. In this paper, we use a subset of 445$\times$750
pixels with 360 spectral bands. The dataset has the spatial resolution of 2 m and spectral width of 5 nm. The dataset contains 16 land cover classes shown in Fig. \ref{fig:Indian10}. The training and test sets used in this study have also been demonstrated in Fig. \ref{fig:Indian10}. Table \ref{tab:SampleIndian10} gives the number of samples, including training and test samples used in the experimental section.

\begin{table}[htbp]
\small\addtolength{\tabcolsep}{-4pt}
  \centering
  \caption{Indian Pines 2010: The number of training samples, test samples, and the total number of samples per class.}
    \begin{tabular}{ccccc}
    \multicolumn{1}{l}{Class No.} & Class Name & \multicolumn{1}{l}{Training Samples} & \multicolumn{1}{l}{Test Samples} & \multicolumn{1}{l}{Samples} \\ \midrule
    1     & Corn-high & 726   & 2661  & 3387 \\
    2     & Corn-mid & 465   & 1275  & 1740 \\
    3     & Corn-low & 66    & 290   & 356 \\
    4     & Soy-bean-high & 324   & 1041  & 1365 \\
    5     & Soy-bean-mid & 2548  & 35317 & 37865 \\
    6     & Soy-bean-low & 1428  & 27782 & 29210 \\
    7     & Residues & 368   & 5427  & 5795 \\
    8     & Wheat & 182   & 3205  & 3387 \\
    9     & Hay   & 1938  & 48107 & 50045 \\
    10    & Grass/Pasture & 496   & 5048  & 5544 \\
    11    & Cover crop 1 & 400   & 2346  & 2746 \\
    12    & Cover crop 2 & 176   & 1988  & 2164 \\
    13    & Woodlands & 1640  & 46919 & 48559 \\
    14    & Highway & 105   & 4758  & 4863 \\
    15    & Local road & 52    & 450   & 502 \\
    16    & Buildings & 40    & 506   & 546 \\ \midrule
          & Total & 10954 & 187120 & 198074 \\
    \end{tabular}%
  \label{tab:SampleIndian10}%
\end{table}%

\subsubsection{Houston University 2013}

\begin{figure} [tbp]\begin{center}
 \includegraphics[width=0.99\linewidth]{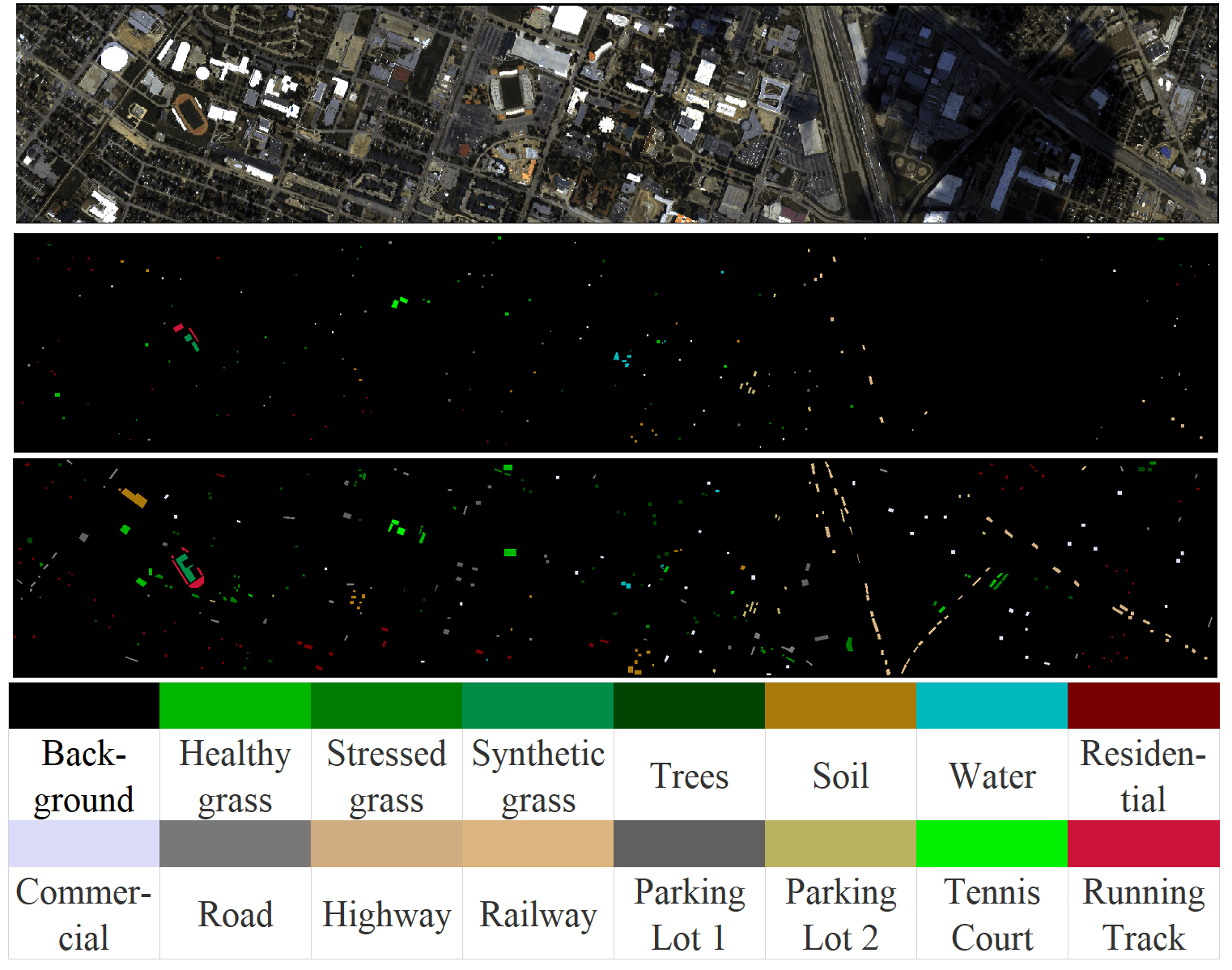}
 \end{center}
 \caption{Houston University 2013 datasets, from top to bottom, RGB composition, Training set, and Test set.}
  \label{fig:HU2013}
 \end{figure}

This dataset was acquired on June 23, 2012, by the Compact Airborne Spectrographic Imager (CASI) over the campus of the University of Houston and the neighboring urban area. The average height of the sensor was 5500ft. The data contain 349 $\times$ 1905 pixels with the spatial resolution of 2.5 m and 144 spectral bands ranging 0.38-1.05 $\mu m$. The dataset includes 15 classes of interests shown in Fig. \ref{fig:HU2013}. A color composite representation of the data and the corresponding training and test samples used in this study are shown in Fig. \ref{fig:HU2013}. The number of training and test samples for different classes of interests used in the experiments are given in Table \ref{tab:Hu2013}.

\begin{table}[htbp]
\small\addtolength{\tabcolsep}{-4pt}
  \centering
  \caption{Houston University 2013: The number of training samples, test samples, and the total number of samples per class.}
    \begin{tabular}{ccccc}
    \multicolumn{1}{c}{Class No.} & \multicolumn{1}{c}{Class Name} & \multicolumn{1}{c}{Training Samples} & \multicolumn{1}{c}{Test Samples} & \multicolumn{1}{c}{Samples} \\\midrule
    \multicolumn{1}{c}{1} & Grass Healthy & \multicolumn{1}{c}{198} & \multicolumn{1}{c}{1053} & \multicolumn{1}{c}{1251} \\
    \multicolumn{1}{c}{2} & \multicolumn{1}{c}{Grass Stressed} & \multicolumn{1}{c}{190} & \multicolumn{1}{c}{1064} & \multicolumn{1}{c}{1254} \\
    \multicolumn{1}{c}{3} & \multicolumn{1}{c}{Grass Synthetic} & \multicolumn{1}{c}{192} & \multicolumn{1}{c}{505} & \multicolumn{1}{c}{697} \\
    \multicolumn{1}{c}{4} & \multicolumn{1}{c}{Tree} & \multicolumn{1}{c}{188} & \multicolumn{1}{c}{1056} & \multicolumn{1}{c}{1244} \\
    \multicolumn{1}{c}{5} & \multicolumn{1}{c}{Soil} & \multicolumn{1}{c}{186} & \multicolumn{1}{c}{1056} & \multicolumn{1}{c}{1242} \\
    \multicolumn{1}{c}{6} & \multicolumn{1}{c}{Water} & \multicolumn{1}{c}{182} & \multicolumn{1}{c}{143} & \multicolumn{1}{c}{325} \\
    \multicolumn{1}{c}{7} & \multicolumn{1}{c}{Residential} & \multicolumn{1}{c}{196} & \multicolumn{1}{c}{1072} & \multicolumn{1}{c}{1268} \\
    \multicolumn{1}{c}{8} & \multicolumn{1}{c}{Commercial} & \multicolumn{1}{c}{191} & \multicolumn{1}{c}{1053} & \multicolumn{1}{c}{1244} \\
    \multicolumn{1}{c}{9} & \multicolumn{1}{c}{Road} & \multicolumn{1}{c}{193} & \multicolumn{1}{c}{1059} & \multicolumn{1}{c}{1252} \\
    \multicolumn{1}{c}{10} & \multicolumn{1}{c}{Highway} & \multicolumn{1}{c}{191} & \multicolumn{1}{c}{1036} & \multicolumn{1}{c}{1227} \\
    \multicolumn{1}{c}{11} & \multicolumn{1}{c}{Railway} & \multicolumn{1}{c}{181} & \multicolumn{1}{c}{1054} & \multicolumn{1}{c}{1235} \\
    \multicolumn{1}{c}{12} & \multicolumn{1}{c}{Parking Lot 1} & \multicolumn{1}{c}{192} & \multicolumn{1}{c}{1041} & \multicolumn{1}{c}{1233} \\
    \multicolumn{1}{c}{13} & \multicolumn{1}{c}{Parking Lot 2} & \multicolumn{1}{c}{184} & \multicolumn{1}{c}{285} & \multicolumn{1}{c}{469} \\
    \multicolumn{1}{c}{14} & \multicolumn{1}{c}{Tennis Court} & \multicolumn{1}{c}{181} & \multicolumn{1}{c}{247} & \multicolumn{1}{c}{428} \\
    \multicolumn{1}{c}{15} & \multicolumn{1}{c}{Running Track} & \multicolumn{1}{c}{187} & \multicolumn{1}{c}{473} & \multicolumn{1}{c}{660} \\\midrule
          & Total & 2832  & 12197 & 15029 \\
    \end{tabular}%
  \label{tab:Hu2013}%
\end{table}%

\subsubsection{Houston University 2018}

\begin{figure} [tbp]\begin{center}
 \includegraphics[width=0.99\linewidth]{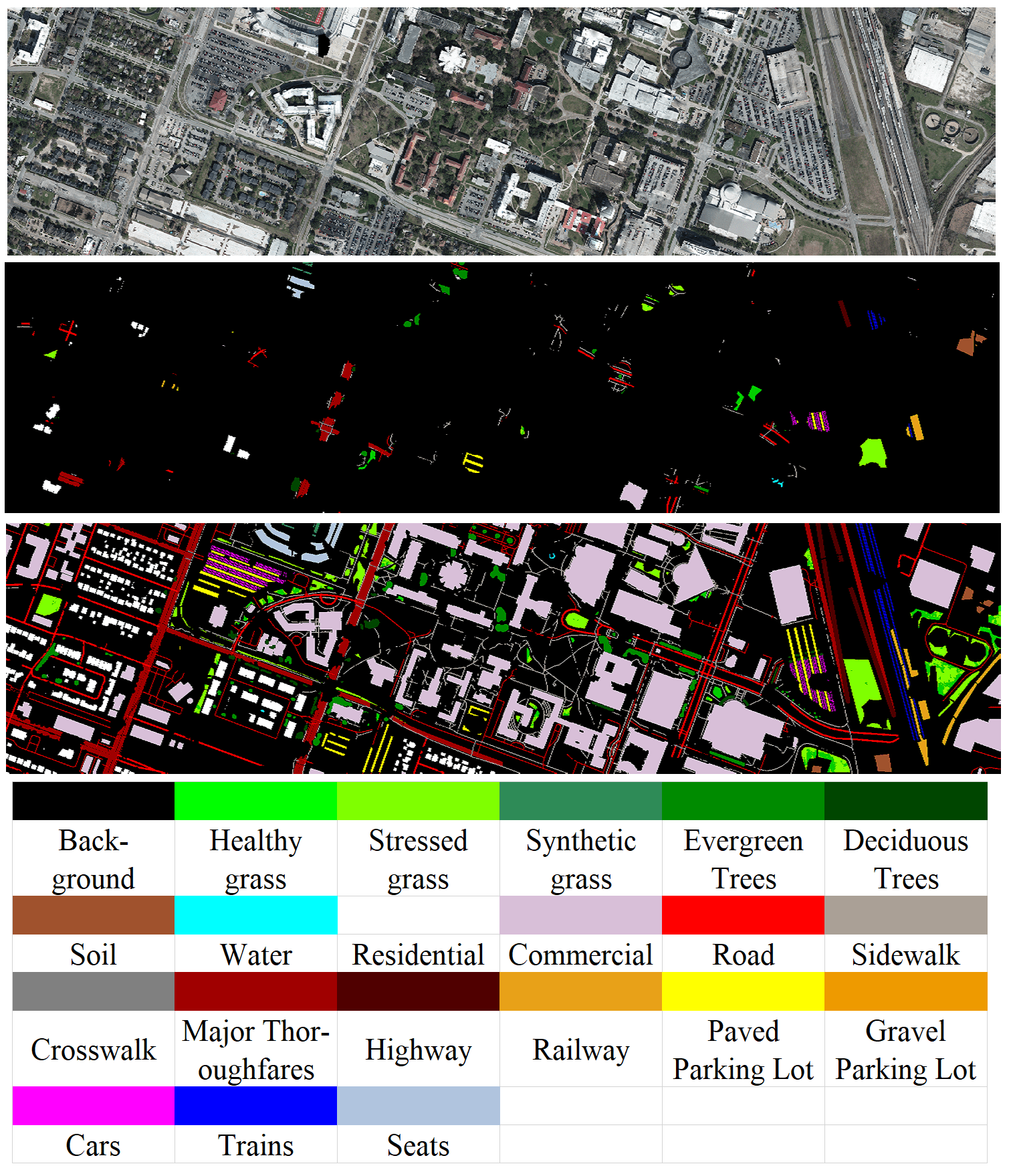}
 \end{center}
 \caption{Houston University 2018 datasets, from top to bottom, The VHR RGB Image (downsampled), Training set, and Test set.}
  \label{fig:HU2018}
 \end{figure}

This dataset was acquired on Feb. 16, 2017 by the hyperspectral imager CASI 1500 over the area of the University of Houston. The data covers the spectral range 380-1050 nm with 48 bands with the ground sampling distance of 1 m. In this article, we utilized the training portion of the whole data set, which was distributed by the Image Analysis and Data Fusion Technical Committee of the IEEE Geoscience and Remote Sensing Society (GRSS) and the University of Houston for the 2018 data fusion contest. It contains 601 $\times$ 2384 pixels and 20 land cover classes of interest shown in Fig.\ref{fig:HU2018}. The VHR RGB image is downsampled and shown in Fig.\ref{fig:HU2018} together with the corresponding training and test samples used in this study. The number of training and test samples for different classes of interest used in the experiments are given in Table \ref{tab:Hu2018}.

\begin{table}[htbp]
\small\addtolength{\tabcolsep}{-4pt}
  \centering
  \caption{Houston University 2018: The number of training samples, test samples, and the total number of samples per class.}
    \begin{tabular}{ccccc}
    \multicolumn{1}{c}{Class No.} & \multicolumn{1}{c}{Class Name} & \multicolumn{1}{c}{Training} & \multicolumn{1}{c}{Test} & \multicolumn{1}{c}{Sample} \\\midrule
    1     & \multicolumn{1}{c}{Healthy grass} & 1458  & 8341  & 9799 \\
    2     & \multicolumn{1}{c}{Stressed grass} & 4316  & 28186 & 32502 \\
    3     & \multicolumn{1}{c}{Synthetic grass} & 331   & 353   & 684 \\
    4     & \multicolumn{1}{c}{Evergreen Trees} & 2005  & 11583 & 13588 \\
    5     & \multicolumn{1}{c}{Deciduous Trees} & 676   & 4372  & 5048 \\
    6     & \multicolumn{1}{c}{Soil} & 1757  & 2759  & 4516 \\
    7     & \multicolumn{1}{c}{Water} & 147   & 119   & 266 \\
    8     & \multicolumn{1}{c}{Residential} & 3809  & 35953 & 39762 \\
    9     & \multicolumn{1}{c}{Commercial} & 2789  & 220895 & 223684 \\
    10    & \multicolumn{1}{c}{Road} & 3188  & 42622 & 45810 \\
    11    & \multicolumn{1}{c}{Sidewalk} & 2699  & 31303 & 34002 \\
    12    & \multicolumn{1}{c}{Crosswalk} & 225   & 1291  & 1516 \\
    13    & \multicolumn{1}{c}{Major Thoroughfares} & 5193  & 41165 & 46358 \\
    14    & \multicolumn{1}{c}{Highway} & 700   & 9149  & 9849 \\
    15    & \multicolumn{1}{c}{Railway} & 1224  & 5713  & 6937 \\
    16    & \multicolumn{1}{c}{Paved Parking Lot } & 1179  & 10296 & 11475 \\
    17    & \multicolumn{1}{c}{Gravel Parking Lot } & 127   & 22    & 149 \\
    18    & \multicolumn{1}{c}{Cars} & 848   & 5730  & 6578 \\
    19    & \multicolumn{1}{c}{Trains} & 493   & 4872  & 5365 \\
    20    & \multicolumn{1}{c}{Seats} & 1313  & 5511  & 6824 \\\midrule
          & Total & 34477 & 470235 & 504712 \\
    \end{tabular}%
  \label{tab:Hu2018}%
\end{table}%

\subsection{Notations}
\label{sec: Notation}
The observed HSI is denoted by ${\bf X} \in\mathbb{R}^{p \times n}$ where $p$ and $n$ are the number of spectral bands and pixels in each band, respectively. $d$ indicates the dimension of the feature space (the subspace). ${\bf X}_m \in\mathbb{R}^{p \times m}$ where $m<n$ denotes the matrix which contains the training samples. ${\bf y}_m \in\mathbb{R}^{1 \times m}$ denotes the vector which contains the class labels where ${y}_{i} \in\left\{1,2,\dots,k\right\}$ and $k$ denotes the number of classes.  
${\bf I}$ is the identity matrix and $\hat{\bf X}$ is the estimate of matrix ${\bf X}$. The Frobenius norm is denoted by $\left\|.\right\|_F$. $\mbox{tr}({\bf X})$ denotes the trace of matrix ${\bf X}$. The definitions of the symbols used in the paper are given in Table \ref{tab:Notation}.
\begin{table*}
\centering
\caption{The different symbols used in this paper and their definition}
\small
\begin{tabular}{l|l}
  \hline
  Symbols & Definition \\ \hline
  $x_i$ & the $i$th entry of the vector ${\bf x}$. \\
  $\bf{X}_{ij}$ & the $(i,j)$th entry of the matrix ${\bf X}$. \\
	${\bf x}_{i}$ & the $i$th column of the matrix ${\bf X}$. \\
	${\bf x}_{(j)}$ & the $j$th row of the matrix ${\bf X}$. \\
  $\left\|{\bf x}\right\|_0$ &  $l_0$-norm of the vector ${\bf x}$ i.e. the number of nonzero entries.\\
  $\left\|{\bf x}\right\|_1$ &  $l_1$-norm of the vector ${\bf x}$, obtained by $\sum_i \left|x_i\right|$.\\
  $\left\|{\bf x}\right\|_2$ &  $l_2$-norm of the vector ${\bf x}$, obtained by $\sqrt{\sum_i x^2_i}$.\\
  $\left\|{\bf X}\right\|_1$ & $l_1$-norm of the matrix ${\bf X}$, obtained by $\sum_{i,j} \left|{\bf X}_{ij}\right|$.\\
  $\left\|{\bf X}\right\|_F$ & Frobenius-norm of the matrix ${\bf X}$, obtained by $\sqrt{\sum_{i,j} {\bf X}^2_{ij}}$.\\
  $\hat{\bf X}$ & the estimate of the matrix ${\bf X}$.\\
  ${\rm tr}({\bf X})$ & the trace of the matrix ${\bf X}$.\\
		$\left\|{\bf X}\right\|_{TV}$ & total variation norm of the matrix ${\bf X}$ is obtained by $\sum_i\mbox{TV}\left({\bf x}_{(i)}\right)$\\
  \hline
\end{tabular}
\label{tab:Notation}
\end{table*}

\section{Shallow Feature Extraction Techniques}
\label{sec:SH_FE}

\subsection{Unsupervised Feature Extraction Techniques}
\label{subsec:UFE}
Unsupervised feature extraction (UFE) often refers to the FE techniques which do not incorporate the knowledge of the ground (ground reference or labeled samples) to extract features. UFE techniques often rely on intrinsic characteristic of the HSI data such as geometric, spatial or spectral information to extract the features. Arguably, the main advantage of UFE compared with the other FE techniques is the lack of need for the training samples, which is of great importance in the case of remote sensing datasets. In this paper, four major UFE groups widely-used for HSI analysis are studied, which are categorized in the following subsections. Fig. \ref{fig:ufe} illustrates the graphical abstracts of those groups. Before explaining the UFE techniques in more details, we briefly refer to three groups of widely-used FE techniques which could also be assumed as UFE, however, will not be studied in details in this paper due to their specific applications. 
The first group includes a range of approaches such as normalized differential vegetation index (NDVI) and normalized differential water index (NDWI) which often rely on the knowledge of the characteristics of the sensors. The second group includes unmixing techniques, which could also be assumed as UFE techniques. They often exploit optimization techniques to show the fractions of materials existing in pixels based on some assumptions on the spectral signatures of the materials. Therefore, the final features extracted represent different materials in the scene at the sub-pixel level \cite{unmixing-review}. The third group includes an impressive number of approaches based on mathematical morphology, which hierarchically extract spatial and contextual information from the input image and usually leads to a significant increase in the number of features \cite{survey-AP}.

\subsubsection{Conventional Data Projection/Transformation Techniques}
\label{subsec:TP}
Numerous UFE techniques fall into this category. The conventional techniques categorized in this group are often designed to linearly project or transform the data, ${\bf X}$, in a lower dimensional feature space (also called subspace) exploiting different non-local intrinsic characteristics of the hyperspectral dataset. The transformation can be given by
\begin{equation}
    {\bf Z}={\bf V}^T{\bf X}
\end{equation}
where ${\bf Z}$ is the projected data in the lower dimensional space and ${\bf V}$ is the transformation matrix or the bases for the subspace. Arguably, principal component analysis (PCA) \cite{PCA} can be considered as the most conventional UFE technique which has been widely used for hyperspectral analysis \cite{PCA_NHAIS}. PCA captures the maximum variance of the signal by projecting the signal on the eigenvectors of the covariance matrix (${\bf C}$) using
\begin{equation}
    \max_{\bf V}  \frac{{\bf V}^T{\bf C}{\bf V}}{{\bf V}^T {\bf V}}.
\end{equation}
A widely used HSI UFE technique is the maximum noise fraction (MNF) \cite{MNF} or noise adjusted principal components (NAPC) \cite{NAPC} which seek a projection in which the signal to noise ratio (SNR) is maximized. MNF uses the following optimization
\begin{equation}
    \max_{\bf V}  \frac{{\bf V}^T{\bf C}{\bf V}}{{\bf V}^T{\bf C}_n {\bf V}},
\end{equation}
where ${\bf C}_n$ is the noise covariance matrix. Another conventional technique is independent component analysis (ICA) \cite{hyvarinenICA}. ICA assumes a linear mixture model of the non-Gaussian independent source signals and the mixing matrix which both are simultaneously estimated and therefore ICA is referred to as blind source separation. ICA has been also widely used for hyperspectral image analysis \cite{villa2}.

To cope with the nonlinearity of the HSI data, the kernel (nonlinear) versions of the aforementioned techniques i.e., kernel MNF \cite{KMNF}, kernel ICA (KICA) \cite{KICA}, and kernel PCA (KPCA) \cite{KPCA} have been also proposed. By using the kernel trick the data are projected into a feature space where the inner product are defined using a kernel function. KICA and KPCA have been used as UFE techniques for change detection and classification in \cite{KICA_ICA_Change} and \cite{MathieuKPCA}, respectively. In \cite{WT_FE}, discrete wavelet transformation (DWT) has been used for hyperspectral feature extraction. Since, DWT does not reduce the dimension, in \cite{WT_FE}, linear discriminant analysis (LDA) has been exploited to reduce the dimension.

\subsubsection{Band Clustering/Splitting and Merging-based Techniques}
\begin{figure*} [tbp]
\centering
\includegraphics[width=1\linewidth]{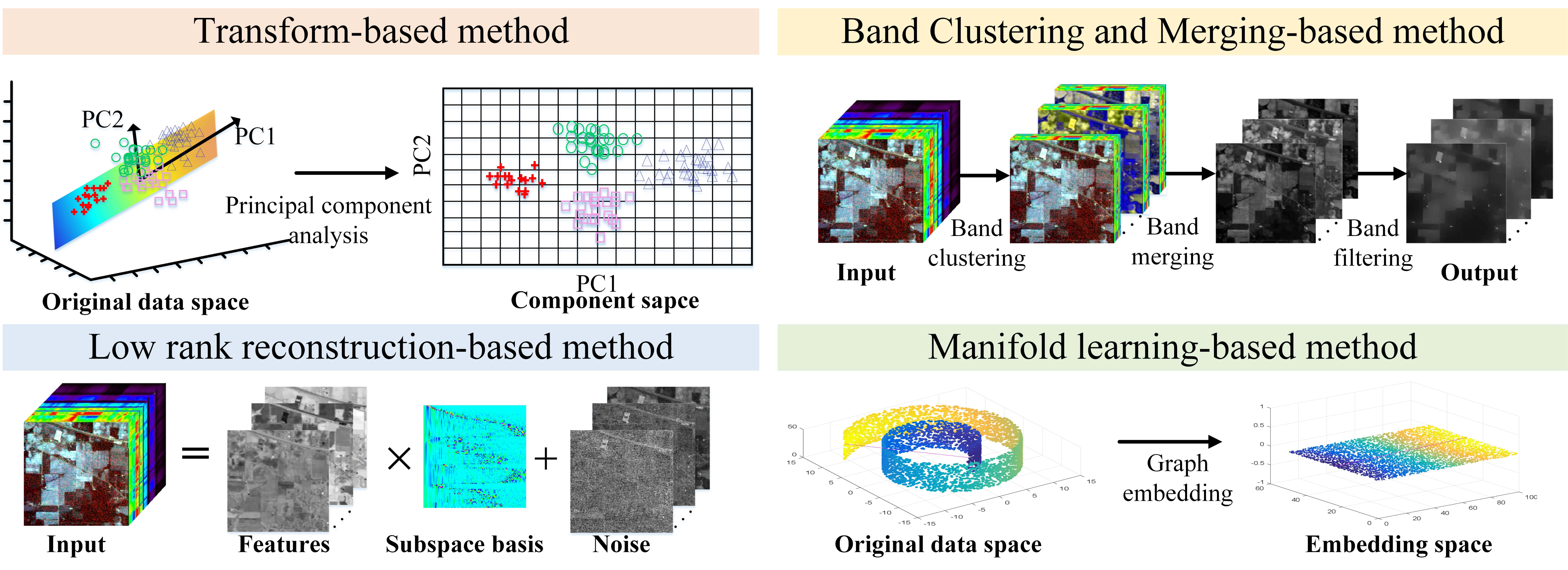}
\caption{Four major categories of unsupervised feature extraction methods.}
 \label{fig:ufe}
\end{figure*}

The top right sub-figure from Fig. \ref{fig:ufe} shows the basic steps of band clustering and merging-based feature extraction methods. As shown in this figure, the core idea behind this group of methods is to split the spectral bands into several groups in which the spectral bands have very high correlation. Hence, the proposed techniques often use similarity and dissimilarity criteria to split the spectral bands into several non-overlapping groups. By selecting or fusing the bands of each group, some representative bands or features of different groups are obtained. Furthermore, followed by the merging step, some band filtering and processing operations can be also used to further improve the discrimination of the resulting features. This group of techniques is often computationally cheap, and thus has been widely used in real applications. On the other hand, spectral information is often neglected by the methods in this category.

For the band clustering and merging two algorithms 
are proposed in \cite{Melba_2001_AV}. 
The first algorithm selects discriminative bases by considering all the classes simultaneously, however, the second algorithm selects the best bases for a pair of classes at a time. In \cite{cluster_BS}, a hierarchical clustering algorithm was introduced to split and cluster the hyperspectral bands where
the representative band for each cluster is selected based on both a mutual information criterion and a divergence-based criterion. Another band clustering technique was proposed in \cite{BandClust} where the splitting is done by minimizing a mutual information criterion applied on averaged bands iteratively. Iterative algorithms were proposed in \cite{Split_Merg}, for both splitting and merging the bands. The splitting procedure is done using the Pearson correlation coefficient between adjacent bands. Then, the merging is applied by averaging over the splited bands.

Besides splitting/clustering and merging hyperspectral bands, another operation is to further improve the feature discrimination by band filtering or processing. For example, a hyperspectral feature extraction using image fusion and recursive filtering was given in \cite{Xudong_FE} where the adjacent bands are fused by averaging and then recursive filtering was used for extracting spatial information. 
In \cite{IID}, the intrinsic image decomposition is applied for processing the merged bands, which can effectively remove information that is not related to the material of different objects. After that, multiple improved versions of intrinsic decomposition-based band processing methods are developed \cite{IID_Gu,SIID_Gu}. In \cite{SE}, a relative total variation-based structure extraction method is applied for band processing, so as to construct multi-scale structural features which are robust to image noise.

\subsubsection{Low-Rank Reconstruction-based Techniques}
\label{subsec:LRRFE}
Low-rank reconstruction-based feature extraction techniques proposed by Rasti \textit{et al.} \cite{RastiPhDThesis, WSRRR,OTVCA,SSLRA} are based on finding an orthogonal subspace by minimizing a constrained cost function. They exploit low-rank models and reconstruction-based optimization frameworks to extract features. The optimization frameworks take into account the prior knowledge of the data using different types of penalties. Due to the noise assumption in the low-rank model used, this group of FE techniques is robust to noise. They are often computationally expensive compared to groups 1 and 2 due to the iterative algorithms used to solve the (non-convex) optimization problem.

 Wavelet-based sparse reduced rank regression (WSRRR) \cite{WSRRR} applies the sparsity prior on the wavelet coefficients considering that the projected data on wavelet bases are sparse. WSRRR uses the model
 \begin{equation}\label{eq: 3Dwavelet}
{\bf X} ={\bf V}^T{\bf Q}{\bf D}_2+ {\bf N},
\end{equation}
where $ {\bf D}_2$ represents 2D wavelet bases, ${\bf X}$ is the observed HSI, ${\bf V}$ contains the orthogonal subspace bases, and ${\bf N}$ is the noise and model error. WSRRR simultaneously estimates the low-rank projection matrix and the wavelet coefficients ${\bf W}$ which minimizes
 \begin{align}\label{eq: WBPCA}\nonumber
	(\hat{\bf V},\hat{\bf Q})=\arg\min _{{\bf Q},{\bf V}}\frac{1}{2}\left\|{\bf X}-{\bf V}^T{\bf Q}{\bf D}_2\right\|^{2}_{F}+&\sum_{j=1}^d\lambda_j\left\|{\bf q}^T_{(j)}\right\|_1\\&~\mbox{s.t.}~ {\bf V}^T{\bf V}={\bf I}.
 \end{align}
Note that the extracted features are given by $\hat{\bf F}=\hat{\bf Q}{\bf D}_2$.

 To capture the spatial (neighboring) information, orthogonal total variation component analysis (OTVCA) has been proposed in \cite{OTVCA} where the HSI is modeled as
 \begin{equation}\label{eq: 3Dwavelet}
{\bf X} = {\bf V}^T{\bf F} + {\bf N},
\end{equation}
where matrix ${\bf F}$ contains the unknown features.
OTVCA assumes that the hyperspectral features are spatially piece-wise smooth and, therefore, exploits the total variation (TV) penalty and simultaneously estimates ${\bf F}$ and ${\bf V}$ using
 \begin{equation}\label{eq: cost}
	{\hat {{\bf F}}}=\arg\min_{{\bf F}}~\frac{1}{2}\left\|{\bf X}-{\bf V}^T{\bf F}\right\|^{2}_{F}+\lambda\sum_{j=1}^d\mbox{TV}\left({\bf f}^T_{(j)}\right),
\end{equation}
 where $$\mbox{TV}({\bf x})=\left\|\sqrt{({\bf D}_h({\bf x}))^2+({\bf D}_v({\bf x}))^2}\right\|_1$$ and ${\bf D}_v$ and ${\bf D}_h$ are the matrix operators to calculate the first order vertical and horizontal differences, respectively, of a vectorized image. 
Recently, sparse and smooth low-rank analysis (SSLRA) was proposed in \cite{SSLRA} which models the HSI based on a combination of sparse and smooth features
 \begin{equation}\label{eq: newmodel}
	{\bf X} = {\bf V}^{T}({\bf F}+{\bf S}) + {\bf N},
\end{equation}
where ${\bf F}$ and ${\bf S}$ contain smooth and sparse features, respectively. SSLRA extracts simultaneously the sparse features, ${\bf S}$, and the smooth ones, ${\bf F}$, by taking into account both sparsity and TV penalties
\begin{align}\label{eq: cost}\nonumber
	({\hat {\bf F}},{\hat {\bf S}},\hat {{\bf V}})=\frac{1}{2}\left\|{\bf X}-{\bf V}^T({\bf F}+{\bf S})\right\|^{2}_{F}+&\lambda_1\left\|{\bf F}\right\|_{\textnormal{TV}}+\lambda_2\left\|{\bf S}\right\|_1  \\ &\mbox{s.t.}~~~ {\bf V}^T{\bf V}={\bf I}.
\end{align}

\subsubsection{Graph Embedding and/or Manifold Learning Techniques}
\label{subsec:GE}
Considering the nonlinear characteristic of HSIs, this group of FE techniques aims to capture the data manifold through local geometric structure of neighboring pixels in the feature space. 
Fig. \ref{fig:ufe} (bottom right) demonstrates the concept of manifold learning FE techniques applied on the Swiss roll dataset. The pink line in the left figure shows the Euclidean distance between two data points in 3D space. It is clear that this line is not an effective metric to measure the similarity of the two points selected in the Swiss roll dataset. On the other hand, after unfolding the dataset which can be represented in 2D space, in the right figure, the Euclidean distance between two data points shown by the pink line is a better representative of the similarity of the two point in the dataset. The FE techniques categorized in this group are indeed designed to capture such a manifold while representing the data in a lower dimensional feature space.

Graph embedding or manifold learning FE techniques often include three main steps, 1) neighborhood pixel selection 2) weight selection, and 3) embedding construction. Isometric mapping (ISOMAP) \cite{ISOMAP1, ISOMAP} is a global geometric nonlinear feature extraction. ISOMAP searches for geodesic distances between data points. It includes three main steps; 1) Constructing a neighborhood graph of the data points. 2) Computing the shortest path
distances between all data points in the neighborhood graph. 3) constructing the lower dimensional embedding vectors which preserves the path distances in the neighborhood graph.
Locally linear embedding (LLE) \cite{LLE},
Laplacian eigenmaps (LE) \cite{EigenMap}, and locality  preserving  projection (LPP) \cite{LPP} are also geometric nonlinear feature extraction based on graph embedding. LLE constructs the embedding graph in three steps; 1) Selecting the neighbors for data points using the $K$ nearest neighbors. 2)  Compute the weights ${\bf A}_{i,j}$ that linearly reconstruct the data points using their neighbors by minimizing the following constrained least-squares
\begin{equation}\label{eq: LLE_W}
    	\min_{{\bf A}}\sum_{i=1}^{n}\left\| {\bf x}_{i}-\sum_{j\in \phi_{i}}{\bf A}_{ij}{\bf x}_{j}\right\|_{2}^{2}~~~{\rm s.t.}~~~\sum_{j\in \phi_{i}}{\bf A}_{ij}={\bf 1},
\end{equation}
where $\phi_{i}({\bf x}_{i})$ contains the neighborhood pixels selected for ${\bf x}_{i}$. We should note that the constrained weights estimated from (\ref{eq: LLE_W}) for every data point are invariant to rotations, rescalings, and translations of that data point and its neighbors and therefore they characterize the intrinsic geometric properties of each neighborhood.
3) Constructing the lower dimensional embedding vectors ${\bf y}$ by minimizing
\begin{align}\nonumber\label{eq: LLE_Y}
    	\min_{{\bf z}}\sum_{i=1}^{n}\left\| {\bf z}_{i}-\sum_{j\in \phi_{i}}{\bf A}_{ij}{\bf z}_{j}\right\|_{2}^{2}~~~{\rm s.t.}~~~&\sum_{i=1}^n{\bf z}_{i}={\bf 0},\\ \frac{1}{n}\sum_{i=1}^n{\bf z}_i{\bf z}_i^T={\bf I}.
\end{align}
We should note that the reconstruction weights ${\bf A}_{ij}$ are fixed in minimization (\ref{eq: LLE_Y}) and therefore the intrinsic geometric properties of the data with dimension $p$ are invariant to such a transformation into a lower dimension $d$.

In \cite{GE}, a general frame-work for graph embedding is given by
\begin{equation}\label{eq6}
    	 \min_{\bf z}\sum_{i,j\in \phi_{i}}\left\| {\bf z}_{i}-{\bf z}_{j}\right\|_{2}^{2}{\bf W}_{ij}~~~ {\rm s.t.}~~~ {\bf Z}{\bf B}{\bf Z}^{\rm T}={\bf I},
\end{equation}
or equivalently
\begin{align}\nonumber
    	\min_{\bf Z}{\rm tr}({\bf Z}({\bf D-W}){\bf Z}^{\rm T})&=\min_{\bf Z}{\rm tr}({\bf Z}{\bf L}{\bf Z}^{\rm T})\\&~~~ {\rm s.t.}~~~ {\bf Z}{\bf B}{\bf Z}^{\rm T}={\bf I},
\end{align}
where ${\bf L}={\bf D-W}$ denotes the Laplacian matrix of the undirected weighted graph $G=\{{ \bf X},{\bf W}\}$ (where ${ \bf X}$ is the vertex set and ${\bf W} \in \mathbb{R}^{n \times n}$ is the similarity matrix), ${\bf D}$ is a diagonal matrix where its entries are given by
\begin{equation}
{\bf D}_{ii}=\sum_{j\ne i}{\bf W}_{ij}, \forall i.
\end{equation}
The diagonal matrix ${\bf B}$ is for the scale
normalization and might also be the Laplacian matrix of
a penalty graph such as $G^p=\{{ \bf X},{\bf W}^p\}$. We should note that the vertices of $G^p$ and $G$ (i.e., ${ \bf X}$) are the same while the similarity matrix (${\bf W}^p$) corresponds to the similarity characteristics suppressed in the lower dimensional feature space (${\bf B}={\bf L}^p={\bf D}^p-{\bf W}^p$, see \cite{GE}).
LLE can be reformulated using graph embedding mentioned above with similarity matrix ${\bf W}_{ij}={\bf A}_{ij}+{\bf A}_{ij}^T-{\bf A}^T_{ij}{\bf A}_{ij}$ if $j\in \phi_i$ otherwise ${\bf W}_{ij}= 0$ and ${\bf B}={\bf I}$ \cite{GE}. ISOMAP, LE, and LPP can be also formulated using graph embedding \cite{GE}. In the viewpoint of graph embedding the main differences between these FE techniques are the selection of the matrices ${\bf W}$ and ${\bf B}$. For instance, LE and LPP use the Gaussian function with the standard deviation $\sigma$ to choose the similarity matrix as
\begin{equation}
\label{eq3}
  {\bf W}_{ij}=
    \begin{cases}
      \exp{\frac{-||{\bf x}_{i}-{\bf x}_{j}||_{2}^{2}}{2\sigma^{2}}},~~~\forall i,j \in \phi_{i}({\bf x}_{i})\\
       0,~~~~~~~~~~~~~~~~~~~\; \text{otherwise.}
    \end{cases}
\end{equation}

We should note that the techniques categorized in this group are assumed as supervised FE methods when they are applied only on the training samples. This is common in the case of HSI due to the large volume of the image which makes the algorithm computationally very expensive. In the following section we will discuss how the ground reference (training samples) can be used to construct the edge matrix, {\bf W}, and therefore those techniques are considered as SFE.

\label{subsec:SFE}
\begin{figure*}[!t]
	  \centering
			\includegraphics[width=0.95\textwidth]{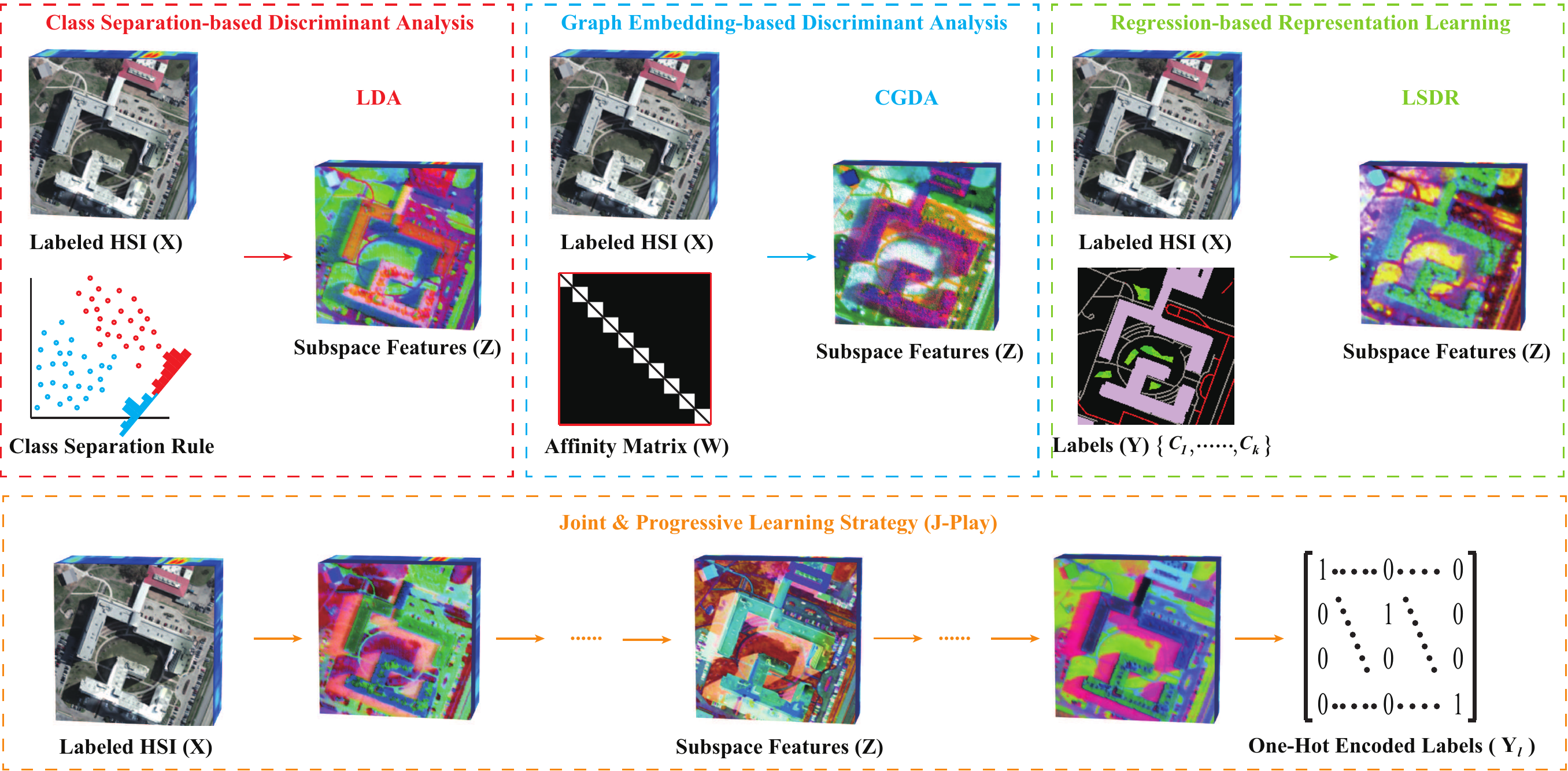}
        \caption{The illustration for supervised FE with four different categories. The obvious differences lie in the use form of label information and learning strategies, i.e., LDA: Fishers rule, CGDA: Affinity matrix, LSDR: Labels, JPlay: Joint use of affinity matrix and one-hot encoded labels.}
\label{fig:Supervised_GRSM}
\end{figure*}
\subsection{Supervised Feature Extraction Techniques}
Unlike unsupervised FE techniques that rely on modeling various prior assumptions of hyperspectral data, supervised methods are capable of extracting class-separable features more effectively, owing to the use of label information. Over the past few decades, some seminal models have been widely developed and applied to perform supervised feature extraction on HSIs, which can be roughly categorized into two streams: \textit{Subspace learning (SL)-based} and \textit{band selection (BS)-based} approaches.

Different from the hand-crafted features \cite{wu2019orsim}, the SL-based approaches learn to extract the low-dimensional representation from the data by formulating different supervised rules in view of label information. There are some typical methods in SL, including linear discriminant analysis (LDA) \cite{LDA}, matrix discriminant analysis (MDA) \cite{hang2016matrix}, decision boundary FE (DBFE) \cite{DBFE}, etc. While the latter, BS, which aims at screening out the representative and informative spectral bands, is unfolded with mutual information-based BS \cite{guo2006band}, rough set, and fuzzy C-means \cite{RoughFCM}, to name a few. To further enhance the class separability, a large number of extended methods have been successfully proposed in recent years, which are subspace LDA (SLDA) \cite{yang2003can}, regularized LDA \cite{bandos2009classification}, local fisher's discriminant analysis (LFDA) \cite{sugiyama2007dimensionality},
feature space discriminative analysis (FSDA) \cite{imani2015feature}, rough-set-based BS \cite{patra2015hyperspectral}, and FE with local spatial modeling \cite{Cao16}.

According to the powerful learning ability of SL methods compared to that of BS-based strategies, we rather focus on reviewing the SL-related FE techniques, in which two main streams -- discriminant analysis FE (DAFE) and regression-induced representation learning (RIRL) -- are emphatically investigated and compared by clarifying their similarities and differences as well as pros and cons, as briefly illustrated in Fig. \ref{fig:Supervised_GRSM}.
\subsubsection{Discriminant Analysis Feature Extraction (DAFE)}
Generally speaking, DAFE seeks to find an optimal projection or transformation matrix ${\bf P} \in \mathbb{R}^{p \times d}$ ($d$ is the dimension of the to-be-estimated subspace) by optimizing certain class-relevant separation criterion associated with the label information. In this process, the estimated subspace ${\bf Z} \in \mathbb{R}^{d \times n}$ that consists of a series of vector ${\bf z}_{i}$ can be obtained by projecting the samples ${\bf X}_{m}=\left\{{\bf x}_{i}\right\}_{i=1}^{m} \in\mathbb{R}^{p \times m}$ onto a decision boundary, which can be generally expressed as ${\bf Z}={\bf P}^{\rm T}{\bf X}$.
Each vector ${\bf z}_{i}$ in ${\bf Z}$ can be collected by ${\bf P}^{\rm T}{\bf x}_{i}$. Depending on the different types of label embedding, DAFE can be subdivided into LDA and its variants, graph-based discriminant analysis (GDA) and its extensions, and kernelized discriminant analysis (KDA).

\textit{$\triangleright$ LDA and Its Variants:}
The traditional LDA linearly transforms the original data into a discriminative subspace by maximizing the Fisher's ratio in the form of generalized Rayleigh quotient, that is, minimizing the intra-class scatter and maximizing inter-class scatter simultaneously. Given a pair-wise training set $\left\{({\bf x}_{1}, {\bf y}_{1}), \dots , ({\bf x}_{i}, {\bf y}_{i}), \dots , ({\bf x}_{m}, {\bf y}_{m})\right\}$,
the objective function of multi-class LDA to estimate the linear projection matrix ${\bf P}$ can be written as follows:
\begin{equation}
\label{SFE_eq1}
    \begin{aligned}
    	 \mathop{\max}_{{\bf P}}\frac{{\rm tr}({\bf P}^{\rm T}{\bf S}_{b}{\bf P})}{{\rm tr}({\bf P}^{\rm T}{\bf S}_{w}{\bf P})},
    \end{aligned}
\end{equation}
where ${\bf S}_{w}$ and ${\bf S}_{b}$ are defined as the within-class scatter matrix and the between-class scatter matrix, respectively. With the constraint of ${\bf P}^{\rm T}{\bf S}_{w}{\bf P}={\bf I}$, the optimization problem in (\ref{SFE_eq1}) can be equivalently converted to one of ${\bf S}_{b}{\bf P}=\lambda{\bf S}_{w}{\bf P}$ by introducing the Lagrange multiplier $\lambda$. The close-form solution to the simplified optimization problem can be deduced by a generalized eigenvalues decomposition (GED).

Due to the sensitivity to complex high-dimensional noises caused by the environmental and instrumental factors and the availability of labeled samples, the original LDA inevitably suffers from an ill-posed statistical degradation, especially in the case of small-scale samples. The degraded reasons mainly lie in the singularity of the two scatter metrics (${\bf S}_{w}$ and ${\bf S}_{b}$), thereby easily leading to the overfitting problem. To improve the stability and generalization, the regularized LDA was proposed by additionally adding a $l_{2}$-norm constraint on ${\bf S}_{w}$ parameterized by $\gamma$ as ${\bf S}_{w}^{\rm reg}={\bf S}_{w}+\gamma{\bf I}$. By replacing ${\bf S}_{w}$ in (\ref{SFE_eq1}) with the regularized ${\bf S}_{w}^{\rm reg}$, the solution in the regularized LDA can be still obtained by the GED solver.

Considering the local neighborhood relations between samples in the process of model learning, LFDA breaks through the bottleneck of those LDA-based methods by assuming that the data are distributed in the nonlinear manifolds rather than a homogeneous Gaussian space. For this purpose, LFDA is capable of effectively excavating the locally underlying structure of the data that lies in the real world. Essentially, LFDA can be regarded as a weighted LDA by locally weighing ${\bf S}_{w}$ and ${\bf S}_{b}$ matrices. Therefore, the two modified scatter matrices, denoted as ${\bf \widetilde{S}}_{w}$ and ${\bf \widetilde{S}}_{b}$, can be formulated as
\begin{equation}
\label{SFE_eq2}
    \begin{aligned}
    	   &{\bf \widetilde{S}}_{w}=\frac{1}{2}\sum_{i=1}^{m}\sum_{j=1}^{m}{\bf W}^{w}_{ij}({\bf x}_{i}-{\bf x}_{j})({\bf x}_{i}-{\bf x}_{j})^{\rm T},\\
    	   &{\bf \widetilde{S}}_{b}=\frac{1}{2}\sum_{i=1}^{m}\sum_{j=1}^{m}{\bf W}^{b}_{ij}({\bf x}_{i}-{\bf x}_{j})({\bf x}_{i}-{\bf x}_{j})^{\rm T},    	
    \end{aligned}
\end{equation}
where the two weights (${\bf W}^{w}$ and ${\bf W}^{b}$) denote the sample-wise similarities. There are several widely-used strategies in calculating such a similarity matrix symbolized by ${\bf W}$. A simple yet effective one is given by ${\bf W}_{ij}=1$, if ${\bf x}_{j} \in \phi_{k}({\bf x}_{i})$, where $\phi_{k}({\bf x}_{i})$ represents the $k$-nearest-neighbor of ${\bf x}_{i}$; otherwise, ${\bf W}_{ij}=0$.
Another commonly-used one was constructed based on the radial basis function (RBF) with a standard derivation of $\sigma$, as defined in (\ref{eq3}).
Please refer to \cite{belkin2003laplacian,he2004locality,zelnik2005self} that might be useful for those who are interested in more types of $\bf W$.
\begin{figure*}[!t]
	  \centering
			\includegraphics[width=0.95\textwidth]{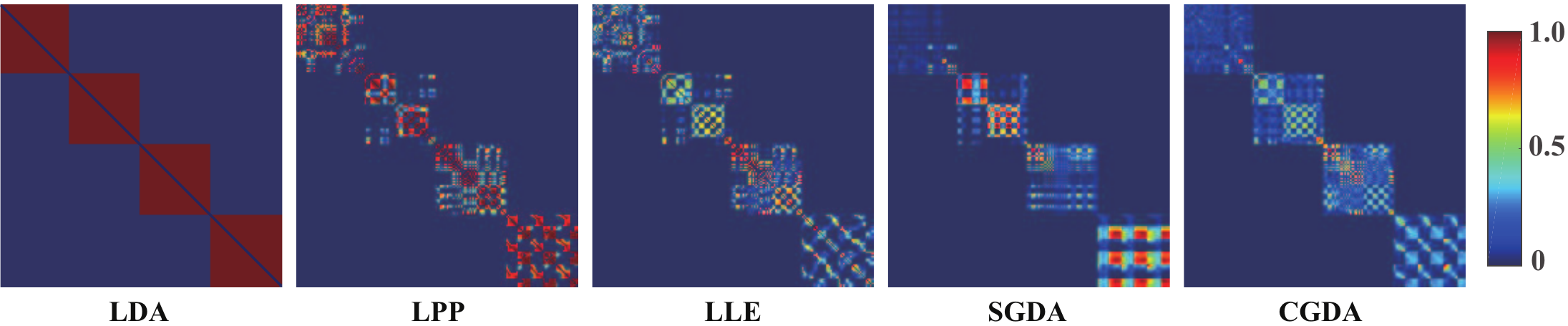}
        \caption{A four-class showcase for affinity matrices (${\bf W}$) with respect to five different approaches, where the connectivity (or edge) of ${\bf W}$ is computed within each class.}
\label{fig:Supervised_graph}
\end{figure*}

Similar to SLDA that first projects the original data into a subspace and then LDA is performed in the transformed subspace, FSDA starts with maximizing the between-spectral scatter matrix (${\bf S}_{f}$) to enhance the differences along the spectral dimension, and similarly, the LDA is further used for extracting the representations of class separability from the feature domain.
In the first step, let $\mu_{i,j}$ be the average value of the $j$-th class and the $i$-th spectral band, then we have the definition of ${\bf S}_{f}$ as follows:
\begin{equation}
\label{SFE_eq3}
    \begin{aligned}
        {\bf S}_{f}=\frac{1}{2}\sum_{i=1}^{p}({\bf h}_{i}-{\bf  \overline{h}})({\bf h}_{i}-{\bf \overline{h}})^{\rm T},
    \end{aligned}
\end{equation}
where ${\bf h}_{i}=[\mu_{i,1}, \mu_{i,2},..., \mu_{i,k}]$ is the spectral representation in the feature space and ${\bf \overline{h}}=\frac{1}{p}\sum_{i=1}^{p}{\bf h}_{i}$. The primary transformation (${\bf P}_{f}$) that aims at improving the spectral discriminant can be estimated with maximizing the trace term of ${\bf S}_{f}$ as
\begin{equation}
\label{SFE_eq4}
    \begin{aligned}
    	 \mathop{\max}_{{\bf P}_{f}}{\rm tr}({\bf P}_{f}^{\rm T}{\bf  S}_{f}{\bf P}_{f}).
    \end{aligned}
\end{equation}
Using the obtained ${\bf P}_{f}$, the latent representation in the feature space ${\bf g}_{i}={\bf P}_{f}^{\rm T}{\bf h}_{i},\; i=1,2,...,p$ can be further fed into the next-step LDA.

\textit{$\triangleright$ GDA and Its Extensions:}
Before revisiting the GDA methods, we first introduce and formulate the general graph embedding (GGE) framework presented in \cite{GE} with Eq. (\ref{eq6}). Obviously, the extracted features ${\bf Z}$ in the GGE framework are determined by the construction of ${\bf W}$ to a great extent. Thus, we will highlight several kinds of representative affinity matrices corresponding to the different graph embedding approaches, i.e. LDA, LE \cite{EigenMap} and its linearized LPP \cite{LPP}, LLE \cite{LLE}, sparse GDA (SGDA) \cite{ly2014sparse}, and collaborative GDA (CGDA) \cite{ly2014collaborative}. Fig. \ref{fig:Supervised_graph} visualizes the affinity matrices given by five different strategies in a four-class case selected from Houston 2013 dataset.

\textbf{LDA-like affinity matrix:} In essence, LDA is vested in a special case of GGE framework with ${\bf D}^{\rm (LDA)}={\bf I}$, whose affinity matrix can be represented as
\begin{equation}
\label{SFE_eq6}
    {\bf W}_{ij}^{\rm (LDA)}=
    \begin{cases}
      \begin{aligned}
      1/N_k, \; \; & \text{if ${\bf x}_{i}$ and ${\bf x}_{j} \in C_{k}$;}\\
      0, \; \; & \text{otherwise,}
      \end{aligned}
    \end{cases}
\end{equation}
where $N_{k}$ is the number of samples belonging to $k$-th class.

\textbf{LPP or LE-based affinity matrix:} One is to be constructed in kernel space with a higher dimension via similarity measurement, i.e. extensively using (\ref{eq3}).

\textbf{LLE-based affinity matrix:} Different from the hand-crafted graph, LLE reconstructs each given sample with its $k$-nearest neighbors by exploiting the linear regression techniques \cite{hong2017learning,hong2019ISPRS}. As a result, the reconstruction coefficients (${\bf A}$) can be obtained by solving the optimization problem of (\ref{eq: LLE_W}).
With the known ${\bf A}$, it is straightforward to derive the needful affinity matrix, denoted as ${\bf W}^{\rm (LLE)}$,
\begin{equation}
\label{SFE_eq8}
    {\bf W}_{ij}^{\rm (LLE)}=
    \begin{cases}
      \begin{aligned}
      &{\bf A}_{ij}+{\bf A}^{\rm T}_{ij}-{\bf A}_{ij}{\bf A}^{\rm T}_{ij}, \; \; \text{if ${\bf x}_{j} \in \phi_{k}({\bf x}_{i})$;}\\
      &0, \qquad \qquad \qquad \qquad \quad \; \text{otherwise,}
      \end{aligned}
    \end{cases}
\end{equation}
thereby inducing the Laplacian matrix as ${\bf L}^{\rm (LLE)}={\bf D}^{\rm (LLE)}-{\bf W}^{\rm (LLE)}=({\bf I}-{\bf A})^{\rm T}({\bf I}-{\bf A})$.

\textbf{SGDA and CGDA-guided affinity matrix:} Similarly to LLE, the affinity matrix can be estimated using the data-driven representation learning, i.e., sparse and collaborative representations \cite{zhang2011sparse,hong2018sulora,hong2019augmented}. Accordingly, the two learning strategies can be equivalent to respectively solving the constrained $l_{1}$-norm optimization problem:
\begin{equation}
\label{SFE_eq9}
    \begin{aligned}
    	 \mathop{\min}_{{\bf W}}\left\|{\bf W}\right\|_{1} \;\; {\rm s.t.} \left\| {\bf X}_{m}{\bf W}-{\bf X}_{m}\right\|_{F}^{2} \leq \epsilon,
    \end{aligned}
\end{equation}
and the $l_{2}$-norm optimization problem:
\begin{equation}
\label{SFE_eq10}
    \begin{aligned}
    	 \mathop{\min}_{{\bf W}}\left\|{\bf W}\right\|_{F}^{2} \;\; {\rm s.t.} \left\| {\bf X}_{m}{\bf W}-{\bf X}_{m}\right\|_{F}^{2} \leq \epsilon.
    \end{aligned}
\end{equation}
The aforementioned affinity matrices can be unified to the GGE framework of (\ref{eq6}).

In addition to SGDA and CGDA (the two baselines), Huang \textit{et al.} \cite{huang2015dimensionality} learned a set of sparse coefficients on manifolds and then preserved the sparse manifold structure in the embedded space. The work in \cite{xue2015simultaneous} extended the existing SGDA to the spatial-spectral graph embedding to address the issues of the spatial variability and spectral multimodality. With the embedding of the intrinsic geometric structure of the data, a Laplacian regularizer CGDA \cite{CGDA} was developed to further improve the graph's confidence. Li \textit{et al.} \cite{li2016sparse} simultaneously integrated the sparsity and low-rankness into the graph for capturing a more robust structure of the data locally and globally. Furthermore, Pan \textit{et al.} \cite{pan2017hyperspectral} further improved the above work by unfolding the HSI data with the form of a tensor.

\textit{$\triangleright$ KDA:}
In reality, the HSI usually exhibits a highly nonlinear data distribution, which may result in difficulties to effectively identify the materials. The solution to this issue is making use of a so-called kernel trick \cite{muller2001introduction} that can map the data of the input space into a new Hilbert space with a higher feature dimension. In the kernel-induced space, the complex nonlinearity of the HSI can be well analyzed in a linearized system. Comparatively, the input to KDA is an inner product of original data pairs, defined as $k({\bf x}_{i}, {\bf x}_{j})$ which can be given by (\ref{eq3}). By introducing the kernel Gram matrix ${\bf K}$ with ${\bf K}_{i,j}={\bf \Phi({\bf x}}_{i})^{\rm T}{\bf \Phi({\bf x}}_{j})=k({\bf x}_{i}, {\bf x}_{j})$, most of previous LDA-based methods can be simply extended to the corresponding kernelized versions, i.e. KLDA and KLFDA can calculate their projections ${\bf P}$ by solving a GED problem of
\begin{equation}
\label{SFE_eq11}
    \begin{aligned}
    	 {\bf K}{\bf L}{\bf K}{\bf P}=\lambda({\bf K}{\bf B}{\bf K}+\gamma{\bf I}){\bf P}.
    \end{aligned}
\end{equation}
Note that ${\bf B}={\bf I}$ in KLDA, while ${\bf L}={\bf L}_{w}$ and ${\bf B}={\bf L}_{b}$ are computed by ${\bf D}_{w}-{\bf W}_{w}$ and ${\bf D}_{b}-{\bf W}_{b}$ in the kernel space, respectively, for KLFDA. Furthermore, for KSGDA and KCGDA, the main difference lies in the computation of the adjacency matrix, which can be performed in the kernel space by solving the general kernel coding problem as follows:
\begin{equation}
\label{SFE_eq12}
    \begin{aligned}
    	 \mathop{\min}_{{\bf W}}{\bf \Omega({\bf W})} \;\; {\rm s.t.} \left\| {\bf \Phi}({\bf X}_{m}){\bf W}-{\bf \Phi}({\bf X}_{m})\right\|_{F}^{2} \leq \epsilon,
    \end{aligned}
\end{equation}
where ${\bf \Omega({\bf W})}$ can be selected to be either sparsity-prompting term $\left\|{\bf W}\right\|_{1}$ of KSGDA or dense (or collaborative) term $\left\|{\bf W}\right\|_{F}^{2}$ of KCGDA. In \cite{gao2010kernel} and \cite{CGDA}, the solutions in (\ref{SFE_eq12}) have been theoretically guaranteed in the same way by solving the problems (\ref{SFE_eq9}) and (\ref{SFE_eq10}) using the alternating direction method of multiplier (ADMM) \cite{ADMM} and least-square regression with Tikhonov regularization \cite{TRLeastSquare}, respectively.
\subsubsection{Regression-induced Representation Learning (RIRL)}
RIRL provides a new insight from the regression's point of view to model the FE behavior by bridging the training samples with the corresponding labels rather than indirectly using the label information in the form of graph or affinity matrix in DAFE-based methods.

\textit{$\triangleright$ Least-Squares Dimension Reduction (LSDR):}
We begin with sliced inverse regression (SIR) \cite{li1991sliced}, which is a landmark in supervised FE techniques. It assumes that the pair-wise data $\{({\bf x}_{i}, {\bf y}_{i})\}_{i=1}^{m}$ are conditionally independent on the to-be-estimated subspace features $\{{\bf z}_{i}\}_{i=1}^{m}$, formulated as $({\bf X}\perp{\bf Y})\mid{\bf Z}$. Following this rule, the LSDR proposed by Suzuki and Sugiyama \cite{LSDR} attempts to find a maximizer of the squared-loss mutual information (SMI) to satisfy the previously mentioned independence assumption. The projections ${\bf P}$ for LSDR can be searched by optimizing the following maximization problem:
\begin{equation}
\label{SFE_eq13}
    \begin{aligned}
    	 \mathop{\max}_{{\bf P}}{\rm SMI}({\bf Z}, {\bf Y}) \;\; {\rm s.t.} \;\; {\bf P}{\bf P}^{\rm T}={\bf I},
    \end{aligned}
\end{equation}
and the SMI to measure a statistical dependence between two discrete variables is defined as
\begin{equation}
\label{SFE_eq14}
    \begin{aligned}
    	 {\rm SMI}({\bf Z}, {\bf Y})=\sum_{{\bf z}\in {\bf Z}}\sum_{{\bf y}\in {\bf Y}} p({\bf z})p({\bf y})\left(\frac{p({\bf z},{\bf y})}{p({\bf z})p({\bf y})}-1 \right)^{2},
    \end{aligned}
\end{equation}
where $p(\bullet)$ is the probability distribution function.

\textit{$\triangleright$ Least-Squares Quadratic Mutual Information (LSQMI):}
Limited by the sensitivity of MI to outliers, authors of \cite{sainui2013direct} designed a more robust LSQMI with the basis of QMI criterion, hence let us define the QMI as
\begin{equation}
\label{SFE_eq15}
    \begin{aligned}
    	 {\rm QMI}({\bf Z}, {\bf Y})=\sum_{{\bf z}\in {\bf Z}}\sum_{{\bf y}\in {\bf Y}}\left(p({\bf z},{\bf y})-p({\bf z})p({\bf y})\right)^{2}.
    \end{aligned}
\end{equation}
Similarly, we solve (\ref{SFE_eq13})-like optimization problem by replacing SMI with QMI.

\textit{$\triangleright$ Least-Squares QMI Derivative (LSQMID):}
Due to the difficulty in accurately computing the derivative of QMI estimator, LSQMI was further extended to a computationally effective LSQMID by estimating the derivative of QMI instead of QMI itself \cite{tangkaratt2017direct}. In this work, authors have demonstrated a more accurate and efficient derivative computation of QMI.

\textit{$\triangleright$ Joint \& Progressive Learning Strategy (JPlay):}
Another MI-free estimation group is latent subspace learning (LSL). One representative LSL performs FE and classification simultaneously in joint learning (JL) fashion \cite{hong2019cospace}. With an expected output ${\bf \Theta}{\bf X}_{m}$, the process can be modeled as
\begin{equation}
\label{SFE_eq16}
    \begin{aligned}
    	 \mathop{\min}_{{\bf P}_k,{\bf \Theta}}\left\|{\bf Y}_{l}-{\bf P}_k{\bf \Theta}{\bf X}_{m}\right\|_{F}^{2}+\frac{\alpha}{2}\left\|{\bf P}_k\right\|_{F}^{2 }\;\; {\rm s.t.} \;\; {\bf \Theta}{\bf \Theta}^{\rm T}={\bf I},
    \end{aligned}
\end{equation}
where ${\bf Y}_{l} \in\mathbb{R}^{k\times m}$ and ${\bf \Theta}\in \mathbb{R}^{d\times p}$ are defined as the one-hot encoded label matrices and the latent subspace projections, respectively. ${\bf P}_k\in\mathbb{R}^{k\times d}$ denotes the regression matrix that connects the learned subspace and the label information. ${\bf Y}_{l} $ can be formulated as
\begin{equation}
\label{SFE_Yl}
    \begin{aligned}
    	 {\bf Y}_{l}=\left[
    	 \begin{matrix}
                1&0&\dots&0&\dots&0\\
    	        0&1&\dots&0&\dots&0\\
    	        &&\dots&\dots&\dots&\\
    	        0&0&\dots&1&\dots&0\\
    	        &&\dots&\dots&\dots&\\
    	        0&0&\dots&0&\dots&1\\
         \end{matrix}
         \right] \;
          \begin{matrix}
                1\\
    	        2\\
    	        \dots\\
    	        j\\
    	        \dots\\
    	        k\\
         \end{matrix}.
    \end{aligned}
\end{equation}
In \cite{hong2019cospace}, the model's solution has been proven to be a closed-form. Moreover, the work in \cite{hong2019learnable} explored a LDA-like graph as a regularizer to learn a spectrally discriminative feature representation, thus (\ref{SFE_eq16}) becomes
\begin{equation}
\label{SFE_eq17}
    \begin{aligned}
    	 \mathop{\min}_{{\bf P}_k,{\bf \Theta}}\left\|{\bf Y}_{l}-{\bf P}_k{\bf \Theta}{\bf X}_{m}\right\|_{F}^{2}+&\frac{\alpha}{2}\left\|{\bf P}_k\right\|_{F}^{2 }+\frac{\beta}{2}{\rm tr}({\bf
    	 \Theta}{\bf X}_{m}{\bf L}{\bf X}_{m}^{\rm T}{\bf \Theta}^{\rm T})\\
    	 &{\rm s.t.} \;{\bf \Theta}{\bf \Theta}^{\rm T}={\bf I}.
    \end{aligned}
\end{equation}

Beyond the JL-based models, Hong \textit{et al.} \cite{JPlay} established a novel multi-layered regression framework by following a joint and progressive learning strategy (JPlay). With the layer-wise auto-reconstruction mechanism effectively against spectral variabilities caused by complex noises and atmospheric effects, the linearized JPlay breaks through the performance bottleneck of traditional linear methods. More specifically, we have the resulting model in the following
\begin{equation}
\label{SFE_eq18}
    \begin{aligned}
    	 \mathop{\min}_{{\bf P}_k,\{{\bf \Theta}_{l}\}_{l=1}^{q}}&\left\|{\bf Y}_{l}-{\bf P}_k{\bf \Theta}_{q}\dots{\bf \Theta}_{1}{\bf X}_{m}\right\|_{F}^{2}+\frac{\alpha}{2}\left\|{\bf P}_k\right\|_{F}^{2 }\\
    	 &+\frac{\beta}{2}\sum_{l=1}^{q}{\rm tr}({\bf
    	 \Theta}_{l}{\bf X}_{l-1}{\bf L}{\bf X}_{l-1}^{\rm T}{\bf \Theta}_{l}^{\rm T})\\
    	 &+\frac{\gamma}{2}\sum_{l=1}^{q}\left\|{\bf X}_{l-1}-{\bf \Theta}_{l}^{\rm T}{\bf \Theta}_{l}{\bf X}_{l-1}\right\|_{F}^{2},\\
    	 & {\rm s.t.} \;{\bf X}_{l}={\bf \Theta}_{l}{\bf X}_{l-1},\;{\bf X}_{l}\geq 0,\;\left\|{\bf x}_{i}\right\|_{2}\leq 1,
    \end{aligned}
\end{equation}
where the soft constraint $\left\|{\bf x}_{i}\right\|_{2}\leq 1$ can be used to relax the orthogonality. It is worth noting that such JL-based strategy can clearly tell the model which features are positive to classification task, owing to the joint strategy of feature extraction and classification.
\begin{figure*}[!t]
	  \centering
		\subfigure{
			\includegraphics[width=0.95\textwidth]{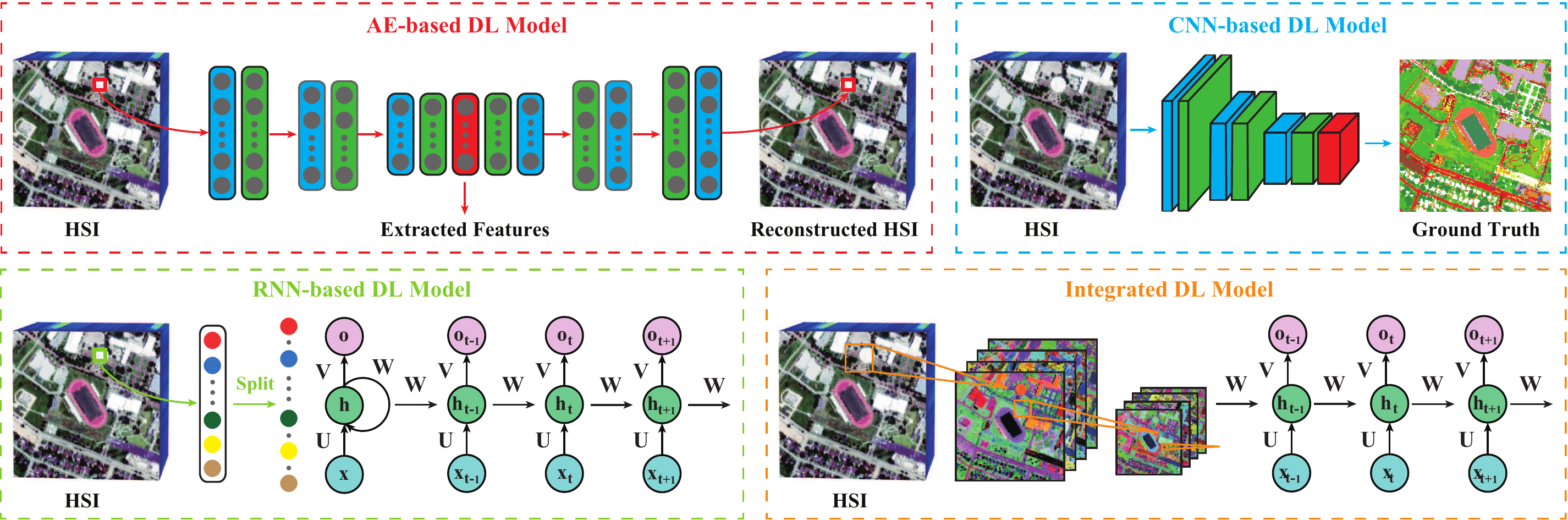}
		}
        \caption{Four major categories of deep learning models.}
\label{fig:DeepFE}
\end{figure*}
\section{Deep Feature Extraction Techniques}
\label{sec:Deep_FE}
Shallow feature extraction techniques often require careful engineering and domain knowledge of experts, which limit their applications. Different from them, deep learning techniques aim at automatically learning high-level features from raw data in a hierarchical fashion. These features are more discriminative, abstract, and robust than shallow ones. Due to their powerful feature representation ability, deep learning techniques have been widely used to extract features from hyperspectral images in recent years \cite{zhang2016deep,8697135}. Among various deep learning models, autoencoders (AEs), convolutional neural networks (CNNs), and recurrent neural networks (RNNs), shown in Fig.$~$\ref{fig:DeepFE}, are the most popular ones. In this section, we will present these models and their applications to hyperspectral feature extraction.

\subsection{AEs}

As demonstrated in Fig.$~$\ref{fig:DeepFE}, AE is mainly comprised of two modules: Encoder and decoder. Encoder maps the input vector $\mathbf{x}$ into a hidden space $\mathbf{h}$, while decoder aims at getting a reconstruction result $\hat{\mathbf{x}}$ of the original input from $\mathbf{h}$. These processes can be formulated as
\begin{equation}
\begin{array}{l}
{\mathbf{h}=f\left(\mathbf{W}_{1} \mathbf{x}+\mathbf{b}_{1}\right)}, \\
{\hat{\mathbf{x}} = f\left(\mathbf{W}_{2} \mathbf{h}+\mathbf{b}_{2}\right)},
\end{array}
\end{equation}
where $\mathbf{W}_{1}$ and $\mathbf{W}_{2}$ denote the weights connecting the input layer to the hidden layer and the hidden layer to the output layer, respectively, $\mathbf{b}_{1}$ and $\mathbf{b}_{2}$ represent the biases of the hidden units and the output units, respectively, and $f$ is a nonlinear activation function. The training of AE is to minimize the residual between $\mathbf{x}$ and $\hat{\mathbf{x}}$. Once trained, the decoder is deleted and the hidden layer $\mathbf{h}$ is considered as a feature representation of $\mathbf{x}$. In order to extract deep features, several AEs are often stacked together, generating a stacked AE (SAE) model. For SAE, the hidden layer in the preceding AE will be used as the input of the subsequent AE.

SAE is perhaps the earliest deep model used to extract features of hyperspectral images \cite{licciardi2018spectral}. One typical benefit of SAEs is that each AE inside the network can be pre-trained using both labeled and unlabeled samples, thus providing better initial values for network parameters compared to the random initialization. After layer-wise pre-training, the fine-tuning of only a few layers can acquire satisfactory discriminant features. This training method is capable of alleviating overfitting problem when there only exist small numbers of training samples in hyperspectral images. In \cite{chen2014}, the spectral information for each pixel was considered as a vector, and fed into an SAE model to extract deep features. These features extracted by SAEs can also be generalized from one image to another image, which was validated in \cite{tao2015} and \cite{kemker2017}.

In order to extract spatial features of each pixel, one often needs to select a local patch or cube centered at the pixel, and then input it into a feature extraction model. Since the inputs of SAEs are vectors, it is difficult to directly process patches or cubes. In \cite{chen2014} and \cite{tao2015}, the local cubes from the first principal components of hyperspectral images were firstly reshaped into vectors, and then fed into SAEs to extract spatial features. In \cite{kang2018} and \cite{deng2019}, Gabor features and extended morphological attribute profiles (i.e., the joint use of shallow and deep feature extraction methods) were used as the inputs of SAEs, making the network easier to extract high-level spatial features. After the extraction of spectral and spatial features, they can be easily concatenated together to generate a spectral-spatial joint feature \cite{chen2014, tao2015}. Compared to the concatenation method, the authors in \cite{kang2018} and \cite{deng2019} proposed to use another SAEs for fusing the spectral and spatial features, which may further enhance the discriminative ability of spectral-spatial features.

Similar to the traditional feature extraction methods, one can also embed some prior or expected information into SAEs. Based upon the assumption that neighboring samples in the input space should have similar hidden representations, graph regularization was added to SAE for preserving such property \cite{ma2016, sun2017}. In \cite{zhou2019}, Zhou \textit{et al.} imposed a local regularization via Fisher discriminant analysis on hidden layers to make the extracted features of samples from the same category close to each other while from different categories as far as possible, thus improving the discriminative ability of SAE. Meanwhile, they also added a diversity regularization term to make SAE extract compact features.

\subsection{Convolutional Neural Networks (CNNs)}
CNNs are the most popularly adopted deep model for hyperspectral feature extraction. As shown in Fig.$~$\ref{fig:DeepFE}, the basic components of a CNN model include convolutional layers, pooling layers, and fully connected layers. The convolutional layers are used to extract features with convolutional kernels (filters), which can be formulated as:
\begin{equation}
\mathbf{X}^{l} = f(\mathbf{X}^{l-1}*\mathbf{W}^{l} + \mathbf{b}^{l}),
\end{equation}
where $\mathbf{X}^{l}$ is the $l$th feature maps, $\mathbf{W}^{l}$ and $\mathbf{b}^{l}$ denote the filters and biases of the $l$th layer, respectively, and `$*$' represents the convolutional operation. After the convolutional layer, the pooling layer is often adopted for reducing the size of the generated feature maps and producing more robust features. On the top of a CNN model, there often exist some fully connected layers, aiming at learning high-level features and outputting the final results of the network.


For hyperspectral images, CNNs can be used for extracting spectral features \cite{hu2015} or spatial features \cite{zhao2016, chen2016,liu2019stfnet}, depending on the inputs of networks. In \cite{hu2015}, Hu \textit{et al.} designed a 1-D CNN model to extract spectral features of each pixel. Compared to the traditional fully-connected networks, CNNs have weight-sharing and local-connection characteristics, making their training processes more efficient and effective. In \cite{zhao2016}, 2-D CNN was explored to extract spatial features from a local cube. Different from SAEs, CNNs do not need to reshape the cube into a vector, thus preserving as much spatial information as possible. However, to make full use of the representation ability of CNNs, two important issues need to be considered. The first issue is the small number of training samples but high-dimensional spectral information, which will easily lead to the overfitting problem. The second issue is the extraction of spectral-spatial joint features, which can improve the classification performance in comparison with using the spectral or spatial feature only.

For the first issue, many widely used strategies in the field of natural image classification, such as dropout and weight decay, can be naturally adopted. In addition, a lot of promising methods have been proposed in the past few years. These methods can be divided into four different classes. The first class of methods is dimensionality reduction. In \cite{zhao2016, chen2016, song2018}, PCA was employed to extract the first principal components of hyperspectral images as inputs of CNNs, thus simplifying the network structures. Similarly, a similarity-based band selection method was used in \cite{ma2018}. However, these dimensionality reduction methods are independent from the following CNNs, which may lose some useful information. Different from them, Ghamisi \textit{et al.} proposed a novel method to adaptively select the most informative bands suitable for the CNN model \cite{Ghamisi2016}. The second class of methods is data augmentation. In \cite{chen2016}, two methods were proposed to generate virtual samples. One is to multiply a random factor and add a random noise to training samples, while the other one is to combine two given samples from the same class with proper ratios. In \cite{Zhan2017}, a data augmentation method based on distance density was proposed. Recently, Kong \textit{et al.} proposed a random zero setting method to generate new samples \cite{Kong2018}.
The third class of methods is transfer learning. In \cite{Yang2017} and \cite{Mei2017}, the authors found that CNNs trained by one hyperspectral data can be transferred to another data acquired by the same sensor, and fine-tuning only a few top layers achieves satisfying results. More interestingly, the works in \cite{Jiao2017, Cheng2018, Liang2018} indicated that CNNs pretrained by natural images can be directly applied to extract spatial features of hyperspectral images. The fourth class of methods is semi-supervised or even unsupervised learning. For examples, Wu and Prasad attempted to use a clustering model to obtain pseudo labels of unlabeled samples, and then combine the training samples and unlabeled samples (with their pseudo labels) together to train their network \cite{wu2018}.

In terms of the second issue, one popularly used method is feeding a local cube, directly cropped from the original hyperspectral image, into a CNN with 3-D convolution kernels for processing the spectral and spatial information simultaneously. The number of channels in the 3-D convolutional kernel is smaller than or equal to that of its input layer. However, the former one dramatically increases the computational complexity due to the simultaneous convolution operators in both spectral domain and spatial domain, while the latter one heavily increases the number of parameters to optimize. Another candidate method is to decouple the task of spectral-spatial feature extraction into two parts: Spectral feature extraction and spatial feature extraction. In \cite{Yang2017} and \cite{xu2018multisource}, a parallel structure was employed to extract spectral-spatial features. Specifically, 1-D CNN and 2-D CNN were designed to extract spectral features and spatial features,respectively; these two features were then concatenated together and fused via a few fully-connected layers. Since 2-D CNN focuses on extracting spatial features, some redundant spectral information can be preprocessed to reduce the computational complexity. In \cite{zhong2018}, a serial structure was also used to extract spectral-spatial features. It firstly applied several $1\times1$ convolutions to extract spectral features and then fed the extracted features into several 3-D convolutions to extract spatial features.

\subsection{Recurrent Neural Networks (RNNs)}
RNNs have been popularly employed to sequential data analysis, such as machine translation and speech recognition. Different from the feedforward neural network, RNN takes advantage of a recurrent edge for connecting the neuron to itself across time. Therefore, it is able to model the probability distribution of sequence data. To make this subsection easier to follow, we first provide a brief and general discussion on RNN. Then, we briefly describe how to use RNN specifically for the classification of hyperspectral images.

Fig.$~$\ref{fig:DeepFE} shows an example of RNN. Given a sequence $\mathbf{x} = (\mathbf{x}_{1}, \mathbf{x}_{2}, \cdots, \mathbf{x}_{T})$, where $\mathbf{x}_{t}, t\in\{1,2,\cdots,T\}$ generally denotes the information at the time $t$, the output of the hidden layer at the $t$th time step is:
\begin{equation}
 \mathbf{h}_{t} = f(\mathbf{U}\mathbf{x}_{t} + \mathbf{W}\mathbf{h}_{t-1} + \mathbf{b}_{h}),
\end{equation}
where $\mathbf{U}$ and $\mathbf{W}$ represent weight matrices from the current input layer to the hidden layer and the preceding hidden layer to the current hidden layer, respectively, $\mathbf{h}_{t-1}$ is the output of the hidden layer at the preceding time, and $\mathbf{b}_{h}$ is a bias vector. According to this equation, it can be observed that the contextual relationships in the time domain are constructed via a recurrent connection. Ideally, $\mathbf{h}_{T}$ will capture most of the information, and can be considered as the final feature of the sequence data. In terms of classification tasks, one often inputs $\mathbf{h}_{t}$ into an output layer $\mathbf{o}_{t}$, which can be described as:
\begin{equation}
 \mathbf{o}_{t} = f(\mathbf{V}\mathbf{h}_{t} + \mathbf{b}_{o}),
\end{equation}
where $\mathbf{V}$ is the weight matrix from the hidden layer to the output layer, and  $\mathbf{b}_{o}$ is a bias vector.


In recent years, RNNs have attracted more and more attention in the field of hyperspectral image feature extraction. To make full use of RNNs, one needs to first ask the following question: How to construct the sequence? An intuitive method is regarding the whole spectral bands as a sequence \cite{CRNN, zhou2019Hyperspectral}. For each pixel, its spectral values are fed into RNN from the first band to the last band, and the output of the hidden layer at the last band is the extracted spectral feature. Different from the traditional sequences in speech recognition or machine translation tasks, the succeeding bands do not depend on the preceding ones. Thus, Liu \textit{et al.} also fed the spectral sequence from the last band to the first band to construct a bidirectional RNN model \cite{CRNN}. Another method is using a local patch or cube to construct the sequence \cite{zhang2018spatial,shi2018multi,zhou2019Hyperspectral}. For examples, Zhou \textit{et al.} regarded the rows of each local patch, cropped from the first principal component of hyperspectral images, as a sequence, and fed them into RNN one by one to extract spatial features \cite{zhou2019Hyperspectral}; Zhang \textit{et al.} adopted each pixel and its neighboring pixels in the cube to form a sequence \cite{zhang2018spatial}. These pixels were firstly sorted according to their similarities to the center pixel, and then fed into RNN sequentially to extract locally spatial features.

In real applications, the constructed sequence may be very long. Take the widely used Indian Pines data as an example, the length of the sequence is 200 (the number of spectral bands) if we use the first method mentioned above to construct the sequence. This sequence will increase the training difficulty, because of the gradients tending to either vanish or explode. In order to deal with this issue, long short-term memory (LSTM) was employed as a more sophisticated recurrent unit \cite{CRNN, ma2019hyperspectral, xu2018spectral, zhou2019Hyperspectral}. The core components of LSTM are three gates: an input gate, a forget gate, and an output gate. These gates together control the flow of information in the network. Similarly, gated recurrent unit (GRU) was also employed, which has only two gates (i.e., an update gate and a reset gate). Compared to LSTM units, GRUs have less numbers of parameters, which may be more suitable for hyperspectral image feature extraction since it usually has a limited number of training samples. Another candidate scheme to address the issue is to divide the long-term sequence into shorter sequences \cite{xu2018spectral, hang2019cascaded}. For example, in \cite{hang2019cascaded}, Hang \textit{et al.} proposed to group the adjacent bands of hyperspectral images into subsequences, and then use RNNs to extract features from them. Since nonadjacent bands have some complementarity, they also used another RNN to fuse the extracted features.

\subsection{Integrated Networks}
In general, AEs and RNNs are good at processing vectorized inputs, thus achieving promising results in terms of spectral feature extraction. However, both of them need to reshape the input patches or cubes into vectors during spatial feature extraction, which may destroy some spatial information. In contrast, CNNs are able to directly deal with image patches and cubes, resulting in more powerful spatial features than AEs and RNNs. It is natural to think whether we can integrate these networks together to make full use of their respective advantages. In the past few years, numerous works have been proposed in this direction.

One kind of integration method is to use each network independently, and then combining their results together \cite{shi2018multi, xu2018spectral, hao2018two, hang2019cascaded}. In \cite{hao2018two}, a parallel framework was proposed to extract spectral-spatial joint features from hyperspectral images. In this framework, SAE was employed to extract spectral features of each pixel, and CNN was used to extract spatial features from the corresponding image patch. These two results were fused by a fully connected layer. Similar to this work, Xu \textit{et al.} also adopted the parallel framework but used LSTM to extract the spectral features \cite{xu2018spectral}. Different from them, Hang \textit{et al.} proposed a serial framework to fuse CNNs and RNNs \cite{hang2019cascaded}. Specifically, they used CNN to extract the spatial features from each band of hyperspectral images, and then used RNN to fuse the extracted spatial features. In \cite{shi2018multi}, Shi and Pun also employed a serial framework to integrate CNN and RNN for spectral-spatial feature extraction.

Another kind of integration method is embedding the core component (i.e., convolutional operators) of CNNs into AEs or RNNs \cite{kemker2017, CRNN}. In \cite{kemker2017}, an unsupervised spectral-spatial feature extraction network was proposed. The whole framework was similar to AEs, which also adopted the so-called encoder-decoder paradigm. However, the fully-connected operators in AEs were replaced by convolutional operators, so that the network can directly extract spectral-spatial joint features from cubes. In \cite{CRNN}, Liu \textit{et al.} proposed a spectral-spatial feature extraction method based on a convolutional LSTM network. Instead of fully connected operators, they also used convolutional operators in LSTM units. For a given cube, each band was fed into the convolutional LSTM unit sequentially. The convolutional operators can extract the spatial features while the recurrent operators can extract the spectral features. The whole network was optimized in an end-to-end way, thus achieving satisfactory performance.

\section{Experimental Results}
\label{sec:ER}
To evaluate the performance of different FE techniques, we selected four techniques from the UFE category (i.e., PCA \cite{PCA}, MSTV \cite{SE}, OTVCA \cite{OTVCA}, LPP \cite{LPP}), four techniques from the SFE category (i.e., LDA \cite{LDA}, CGDA \cite{CGDA}, LSDR \cite{LSDR}, and JPlay \cite{JPlay}), and five techniques from the deep FE category (i.e., SAE \cite{SAE}, RNN \cite{hang2019cascaded}, CNN \cite{CNN}, CAE \cite{CAE}, and CRNN \cite{CRNN}). Here, we set the tuning parameters for those algorithms before representing the experimental results.

\subsection{Algorithm Setup}
The parameter setting usually plays a crucial role in assessing the performance of FE algorithms. Subspace dimension (or number of features, $d$) is a common parameter for all compared algorithms. Selection of the number of features is a hard task for hyperspectral image analysis. The endmember selection/extraction, subspace identification, and/or rank selection are all refer to this issue \cite{Ghamisi-review-2017,HySURE,HySime}. For a fair and simplified comparison, the parameter $d$ is unifiedly assigned to the same value, which is equal to the number of classes ($k$). We should note that $d$ in LDA is automatically determined as $k-1$, due to the class separability (fisher's criterion).

\subsubsection{Unsupervised FE}
\begin{itemize}
\item PCA: This method is a parameter free technique.
\item MSTV: In \cite{SE}, all parameters are adjusted using a trial and error approach. The multi-scale parameters adjusting the degree of smoothness (as suggested in \cite{SE}) are set to 0.003, 0.02, and 0.01. The spatial scale for the structure extraction in three levels (as suggested in \cite{SE}) is set to 2, 1, and 3.
 \item OTVCA: This method is initialized as recommended in \cite{OTVCA}. The tuning parameter $\lambda$ which controls the level of smoothness applied on the features is set to one percent of the maximum intensity range of the datatsets.
\item LPP: The number of neighbors is set to 12. The bandwidth of the Gaussian kernel is set to 1.

\end{itemize}
\subsubsection{Supervised FE}

A common strategy for a model selection is to run cross-validation (CV) on the training set, since the labeled samples are available in SFE. Therefore, we used the CV strategy on the following studied algorithms for parameter selection.
\begin{itemize}
    \item LDA: This method can be viewed as a baseline for SFE. There is no additional parameter in LDA. 
    \item CGDA: The Eq. (23) can be tuned to a regularized optimization problem, where one extra parameter -- regularized $l_{2}$-norm -- needs to be set in advance in the process of graph construction, which can be searched in the range of $\{10^{-4}, 10^{-3}, 10^{-2}, \dots, 10^{1}, 10^{2}\}$ by CV. In the experiments, $0.1$ is used for all three datasets.
    \item LSDR: Two parameters are involved in LSDR; The standard deviation for the Gaussian function and the regularization parameter which are selected in the range of $\{0.05,0.1,\dots,0.95,1\}$ and $\{10^{-2}, 10^{-1} \dots, 10^{1}, 10^{2}\}$, respectively, using CV. Finally, $\sigma$ and $\lambda$ are both set to $1$ in our experiments.
    \item JPlay: There are three regularization parameters ($\alpha$, $\beta$, and $\gamma$) that need to be set in the JPlay model (\ref{SFE_eq18}). With the CV conducted on the training set of three different datasets, the regularization parameters are selected in the ranges of $\{10^{-2}, 10^{-1} \dots, 10^{1}, 10^{2}\}$, yielding the final setting of $(\alpha,\beta,\gamma)=(0.1,1,1)$ for the first dataset, $(\alpha,\beta,\gamma)=(0.1,0.1,1)$ for the second dataset, and $(\alpha,\beta,\gamma)=(0.1,1,1)$ for the last dataset.
\end{itemize}

\subsubsection{Deep FE}
\begin{itemize}
\item SAE: The input of SAE is the original spectral information of each pixel. Three hidden layers are used. The numbers of neurons from the first to the third hidden layer are set to 32, 64, and 128, respectively. ReLU is adopted as the activation function for each hidden layer.
\item CNN: The input of CNN is a small cube with a size of $16\times16\times p$, where $p$ represents the number of spectral bands for each data. Three convolutional layers are used. Each convolutional layer is sequentially followed by a batch normalization layer, a ReLU activation function, and a max-pooling layer. Note that the last pooling layer is an adaptive max-pooling layer, making the output size equals to $1\times1$ for any input sizes. The kernel size for each convolution is $3\times3$, and the numbers of kernels from the first to the third convolutions are set to 32, 64, and 128, respectively. Padding operators are used to preserve the spatial size after each convolutional operator.
\item PCNN: PCA is applied prior to CNN to reduce the spectral dimension of the
HSI. The number of reduced dimension by PCA is set to the number of classes ($k$). The input cube for CNN is of size $16\times16\times k$. 
\item RNN: The input of RNN is the same as the input of SAE. Two recurrent layers with GRU are employed. The number of neurons in each recurrent layer is set to 128.
\item Integrated Networks: Convolutional AE (CAE) and convolutional RNN (CRNN) are selected as two representative integrated networks. The input for them is the same as that for the CNN. For CAE, three convolutional layers and three de-convolutional layers are adopted. All of them use $3\times3$ kernels. The numbers of kernels from the first to the third convolutional layers are set to 32, 64, and 128, respectively. In contrast, the numbers of kernels from the first and the third de-convolutional layers are set to 64, 32, and $p$, respectively. Similar to \cite{CRNN}, CRNN adopts two recurrent layers with convolutional LSTM units. For both recurrent layers, $3\times3$ convolutional kernels are applied. The numbers of kernels for the first and the second recurrent layers are set to 32 and 64, respectively.
\end{itemize}

All of the above deep learning related models are implemented in the PyTorch framework. In order to optimize them, we use the Adam algorithm with default parameters. The batch size, the learning rate, and the number of training epochs are set to 128, 0.001, and 200, respectively. To reduce the effects of random initialization, all of the deep learning models are repeated 5 times, and the mean values are reported. 

\subsubsection{Random Forest Classifier}
Apart from the deep FE technqiues, all the other FE techniques use random forest (RF) to perform the classification task. The number of trees selected for RF is set to 200. We approximately set the number of the prediction variable to the square root of the number of input bands.

\begin{table*}[htbp]
  \centering
  \caption{Classification accuracies obtained on features extracted from the Indian Pines 2010 dataset using different shallow and deep feature extraction techniques. }
  \resizebox{1\textwidth}{!}{
    \begin{tabular}{c|c|cccc|cccc|cccccc}
    \toprule
    \multicolumn{15}{c}{Indian Pines 2010} \\
    \midrule
          &       & \multicolumn{8}{c|}{Shallow FE}                               & \multicolumn{6}{c}{Deep FE} \\
    \midrule
          &       & \multicolumn{4}{c|}{UFE}      & \multicolumn{4}{c|}{SFE}      & \multicolumn{6}{c}{SFE} \\
    \midrule
          & Spectral & PCA   & MSTV  & OTVCA & LPP   & LDA   & CGDA  & LSDR  & JPlay & SAE   & RNN   & CNN   & CAE  & CRNN  & PCNN \\
    \midrule
    1     & 0.9260 & 0.9064 & \textbf{0.9992} & 0.9260 & 0.8850 & 0.9628 & 0.8144 & 0.8459 & 0.9230 & 0.9327 & 0.8829 & 0.9275 & 0.9432 & 0.9590 & 0.9397 \\

    2     & 0.8769 & 0.9976 & \textbf{1.0000} & \textbf{1.0000} & 0.9976 & \textbf{1.0000} & 0.9961 & 0.8933 & 0.9984 & 0.9573 & 0.9178 & 0.9544 & 0.9961 & 0.8596 & 0.9998 \\

    3     & 0.8862 & 0.9724 & 0.9724 & 0.9897 & 0.9724 & 0.9724 & 0.9759 & 0.9655 & 0.9862 & 0.9683 & 0.9405 & 0.9890 & \textbf{0.9986} & 0.8241 & 0.9938 \\

    4     & 0.6888 & 0.7762 & \textbf{1.0000} & 0.8953 & 0.8742 & 0.9270 & 0.7474 & 0.8194 & 0.8300 & 0.7508 & 0.7621 & 0.8840 & 0.8822 & 0.8569 & 0.8930 \\

    5     & 0.8058 & \textbf{0.8855} & 0.8039 & 0.8151 & 0.8682 & 0.8802 & 0.8096 & 0.8394 & 0.8665 & 0.8488 & 0.8474 & 0.8692 & 0.8570 & 0.8638 & 0.8706 \\

    6     & 0.8172 & 0.8797 & \textbf{0.9946} & 0.7094 & 0.9739 & 0.7883 & 0.8284 & 0.9397 & 0.9418 & 0.9013 & 0.9204 & 0.9268 & 0.9167 & 0.6975 & 0.9127 \\

    7     & 0.4170 & 0.5954 & 0.6792 & \textbf{0.7166} & 0.6985 & \textbf{0.7166} & 0.6845 & 0.7030 & 0.6958 & 0.6265 & 0.5795 & 0.6507 & 0.6818 & 0.6348 & 0.7103 \\

    8     & 0.2530 & 0.2583 & 0.2599 & 0.2768 & 0.2961 & 0.2955 & 0.2431 & 0.2952 & 0.2758 & 0.5349 & 0.2840 & 0.8776 & 0.6776 & \textbf{0.9934} & 0.4725 \\

    9     & 0.6545 & 0.7498 & 0.7048 & 0.7943 & \textbf{0.8913} & 0.8452 & 0.7971 & 0.8419 & 0.8142 & 0.8732 & 0.8533 & 0.8302 & 0.8336 & 0.8194 & 0.8621 \\

    10    & 0.8229 & 0.9406 & 0.9594 & 0.9368 & 0.9019 & \textbf{0.9804} & 0.8514 & 0.7761 & 0.9663 & 0.9015 & 0.8096 & 0.8368 & 0.8752 & 0.7946 & 0.9289 \\

    11    & 0.6658 & 0.8402 & 0.9224 & \textbf{0.9945} & 0.8943 & 0.9288 & 0.7195 & 0.7651 & 0.8052 & 0.8121 & 0.7414 & 0.7440 & 0.7633 & 0.6492 & 0.9165 \\

    12    & 0.9985 & \textbf{1.0000} & \textbf{1.0000} & \textbf{1.0000} & 0.9995 & \textbf{1.0000} & \textbf{1.0000} & \textbf{1.0000} & \textbf{1.0000} & 0.9998 & 0.9765 & 0.9945 & 0.9838 & 0.9748 & 0.9991 \\

    13    & 0.9468 & \textbf{0.9962} & 0.9879 & 0.9888 & 0.9925 & 0.9830 & 0.9580 & 0.9738 & 0.9819 & 0.9621 & 0.9427 & 0.9925 & 0.9930 & 0.9892 & 0.9959 \\

    14    & 0.8783 & 0.9000 & 0.9615 & 0.9145 & 0.9344 & 0.9174 & 0.8756 & 0.8628 & 0.8953 & 0.9094 & 0.9030 & 0.9984 & 0.9985 & 0.8981 & \textbf{0.9993} \\

    15    & 0.9333 & 0.9667 & 0.9511 & 0.9933 & 0.9489 & 0.9756 & 0.9333 & 0.9311 & 0.9600 & 0.9307 & 0.8119 & 0.9947 & 0.9942 & 0.7556 & \textbf{0.9978} \\

    16    & 0.3735 & 0.2036 & 0.2885 & 0.1601 & 0.5217 & 0.1719 & 0.3439 & 0.4901 & 0.4466 & 0.2053 & 0.2060 & 0.0980 & 0.1700 & \textbf{0.6028} & 0.1051 \\
    \midrule
    AA    & 0.7465 & 0.8043 & 0.8428 & 0.8194 & \textbf{0.8532} & 0.8341 & 0.7861 & 0.8089 & 0.8367 & 0.8197 & 0.7737 & 0.8480 & 0.8478 & 0.8233 & 0.8498 \\

    OA    & 0.7866 & 0.8598 & 0.8561 & 0.8378 & \textbf{0.9112} & 0.8748 & 0.8370 & 0.8748 & 0.8829 & 0.8836 & 0.8655 & 0.8945 & 0.8911 & 0.8525 & 0.9018 \\

    $\kappa$  & 0.7390 & 0.8297 & 0.8247 & 0.8054 & \textbf{0.8909} & 0.8481 & 0.8010 & 0.8466 & 0.8571 & 0.8580 & 0.8355 & 0.8716 & 0.8673 & 0.8213 & 0.8802 \\
    \bottomrule
    \end{tabular}%
    }
  \label{tab:IP2010}%
\end{table*}%

\begin{table*}[htbp]
  \centering
  \caption{Classification accuracies obtained on features extracted from the Houston university 2013 dataset using different shallow and deep feature extraction techniques.}
  \resizebox{1\textwidth}{!}{
    \begin{tabular}{c|c|cccc|cccc|cccccc}
    \toprule
    \multicolumn{15}{c}{Houston 2013} \\
    \midrule
          &       & \multicolumn{8}{c|}{Shallow FE}                               & \multicolumn{6}{c}{Deep FE} \\
    \midrule
          &       & \multicolumn{4}{c|}{UFE}      & \multicolumn{4}{c|}{SFE}      & \multicolumn{6}{c}{SFE} \\
    \midrule
          & Spectral & PCA & MSTV  & OTVCA & LPP   & LDA   & CGDA  & LSDR  & JPlay & SAE   & RNN   & CNN   & CAE  & CRNN  & PCNN \\
    \midrule
    1     & 0.8262 & \textbf{0.8272} & 0.8025 & 0.8205 & 0.8110 & 0.8177 & 0.8139 & 0.8120 & 0.7768 & 0.8217 & 0.8182 & 0.8104 & 0.8154 & 0.8245 & 0.8089 \\

    2     & 0.8318 & 0.8393 & 0.8412 & 0.8515 & 0.8214 & 0.8355 & 0.8327 & 0.8553 & \textbf{0.9662} & 0.8274 & 0.8153 & 0.8425 & 0.8167 & 0.8412 & 0.8293 \\

    3     & 0.9782 & \textbf{1.0000} & 0.9822 & \textbf{1.0000} & \textbf{1.0000} & \textbf{1.0000} & \textbf{1.0000} & \textbf{1.0000} & 0.9980 & 0.9895 & 0.9939 & 0.8594 & 0.7731 & 0.9156 & 0.8432 \\

    4     & 0.9138 & 0.9081 & 0.7633 & 0.8873 & 0.9479 & 0.8920 & 0.9053 & 0.8864 & 0.9564 & \textbf{0.9773} & 0.9040 & 0.9170 & 0.9153 & 0.9129 & 0.9159 \\

    5     & 0.9659 & 0.9886 & 0.9915 & \textbf{0.9991} & 0.9867 & 0.9384 & 0.9915 & 0.9688 & 0.9782 & 0.9438 & 0.9389 & 0.9699 & 0.9585 & 0.9881 & 0.9824 \\

    6     & 0.9930 & 0.9930 & 0.9580 & 0.9580 & 0.9790 & \textbf{1.0000} & 0.8741 & 0.9860 & 0.9930 & 0.9874 & 0.9678 & 0.8769 & 0.9776 & 0.9483 & 0.9497 \\

    7     & 0.7463 & 0.8927 & 0.6362 & 0.7090 & \textbf{0.9123} & 0.7901 & 0.8535 & 0.8526 & 0.7817 & 0.7293 & 0.7392 & 0.8802 & 0.8694 & 0.8642 & 0.8627 \\

    8     & 0.3305 & 0.4606 & 0.5992 & 0.6724 & 0.4311 & 0.7379 & 0.4302 & 0.4710 & 0.7806 & 0.3792 & 0.4153 & 0.6344 & 0.6762 & 0.5305 & \textbf{0.8351} \\

    9     & 0.6771 & 0.7885 & 0.8706 & \textbf{0.9008} & 0.7413 & 0.6449 & 0.7186 & 0.6752 & 0.7592 & 0.7145 & 0.7367 & 0.8595 & 0.8540 & 0.8404 & 0.8691 \\

    10    & 0.4295 & 0.4749 & 0.6612 & \textbf{0.8398} & 0.4595 & 0.4662 & 0.4826 & 0.5792 & 0.6014 & 0.5556 & 0.5373 & 0.5674 & 0.5782 & 0.4514 & 0.6168 \\

    11    & 0.7011 & 0.7268 & 0.9820 & \textbf{0.9924} & 0.7306 & 0.7239 & 0.7287 & 0.5806 & 0.6983 & 0.6231 & 0.7250 & 0.7417 & 0.7292 & 0.6186 & 0.7913 \\

    12    & 0.5485 & 0.9145 & 0.7349 & \textbf{0.9625} & 0.7560 & 0.6513 & 0.7656 & 0.5687 & 0.7858 & 0.6305 & 0.7606 & 0.9379 & 0.9402 & 0.8440 & 0.9593 \\

    13    & 0.6140 & 0.7754 & 0.6982 & 0.7789 & 0.8105 & 0.6105 & 0.7719 & 0.6702 & 0.7509 & 0.4516 & 0.6656 & 0.8835 & \textbf{0.8968} & 0.8414 & 0.8765 \\

    14    & 0.9838 & 0.9919 & \textbf{1.0000} & \textbf{1.0000} & 0.9960 & 0.9919 & 0.9879 & 0.9595 & 0.9879 & 0.9692 & 0.9850 & 0.9943 & 0.9773 & 0.9603 & 0.9968 \\

    15    & 0.9789 & 0.9746 & \textbf{1.0000} & 0.9789 & 0.9746 & 0.9831 & 0.9852 & 0.9514 & 0.9831 & 0.9732 & 0.9607 & 0.8072 & 0.7471 & 0.9345 & 0.8592 \\
    \midrule
    AA    & 0.7679 & 0.8371 & 0.8347 & \textbf{0.8901} & 0.8239 & 0.8056 & 0.8094 & 0.7878 & 0.8532 & 0.7716 & 0.7976 & 0.8388 & 0.8350 & 0.8210 & 0.8664 \\

    OA    & 0.7278 & 0.8058 & 0.8088 & \textbf{0.8753} & 0.7874 & 0.7745 & 0.7789 & 0.7524 & 0.8280 & 0.7436 & 0.7646 & 0.8239 & 0.8184 & 0.7921 & 0.8526 \\

     $\kappa$  & 0.7076 & 0.7895 & 0.7923 & \textbf{0.8648} & 0.7700 & 0.7552 & 0.7604 & 0.7315 & 0.8134 & 0.7235 & 0.7469 & 0.8096 & 0.8036 & 0.7761 & 0.8404 \\
    \bottomrule
    \end{tabular}%
    }
  \label{tab:HU2012}%
\end{table*}%

\begin{table*}[htbp]
  \centering
  \caption{Classification accuracies obtained on features extracted from the Houston university 2018 dataset using different shallow and deep feature extraction techniques.}
  \resizebox{1\textwidth}{!}{
    \begin{tabular}{c|c|cccc|cccc|cccccc}
    \toprule
    \multicolumn{15}{c}{Houston 2018} \\
    \midrule
          &       & \multicolumn{8}{c|}{Shallow FE}                               & \multicolumn{6}{c}{Deep FE} \\
    \midrule
          &       & \multicolumn{4}{c|}{UFE}      & \multicolumn{4}{c|}{SFE}      & \multicolumn{6}{c}{SFE} \\
    \midrule
          & Spectral & PCA & MSTV  & OTVCA & LPP   & LDA   & CGDA  & LSDR  & JPlay & SAE   & RNN   & CNN   & CAE  & CRNN  & PCNN \\
    \midrule
        1     & 0.3088 & \textbf{0.8781} & 0.0536 & 0.6842 & 0.6618 & 0.6256 & 0.7575 & 0.7969 & 0.5991 & 0.7940 & 0.5702 & 0.7516 & 0.4428 & 0.6338 & 0.6638 \\

    2     & 0.7603 & 0.8396 & 0.7046 & 0.6376 & 0.8122 & 0.8474 & 0.8076 & 0.7747 & 0.8347 & 0.7893 & 0.6975 & 0.8173 & \textbf{0.8849} & 0.8707 & 0.8376 \\

    3     & \textbf{1.0000} & \textbf{1.0000} & \textbf{1.0000} & 0.9972 & \textbf{1.0000} & \textbf{1.0000} & \textbf{1.0000} & \textbf{1.0000} & \textbf{1.0000} & \textbf{1.0000} & 0.9972 & 0.7739 & 0.8482 & 0.9924 & 0.8045 \\

    4     & 0.9134 & 0.9494 & 0.6238 & 0.6775 & 0.9453 & 0.9059 & 0.9265 & 0.9276 & 0.9298 & 0.9221 & 0.8613 & 0.9444 & 0.9362 & 0.9439 & \textbf{0.9595} \\

    5     & 0.4119 & 0.4668 & 0.2676 & 0.1679 & 0.4728 & 0.5258 & 0.4661 & 0.4289 & 0.3971 & 0.4982 & 0.4040 & 0.4330 & 0.5396 & \textbf{0.5404} & 0.4800 \\

    6     & 0.2570 & 0.2990 & \textbf{0.3835} & 0.3164 & 0.3008 & 0.2910 & 0.2776 & 0.2726 & 0.2780 & 0.2585 & 0.2537 & 0.3050 & 0.3080 & 0.2902 & 0.3377 \\

    7     & 0.3025 & 0.3025 & 0.3025 & 0.3025 & 0.3025 & 0.3025 & 0.3109 & 0.3025 & 0.2857 & 0.3025 & 0.3025 & 0.2908 & 0.2723 & 0.2997 & \textbf{0.3176} \\

    8     & 0.7657 & 0.7675 & 0.7599 & 0.7417 & 0.7785 & 0.7849 & 0.7544 & 0.7518 & 0.7771 & 0.7216 & 0.7356 & 0.8538 & 0.8583 & 0.8092 & \textbf{0.8677} \\

    9     & 0.3849 & 0.3877 & 0.5767 & 0.5990 & 0.4887 & 0.3917 & 0.3672 & 0.5255 & 0.5877 & 0.6302 & 0.4186 & 0.7970 & 0.7371 & 0.3717 & \textbf{0.8659} \\

    10    & 0.3603 & 0.4360 & 0.3747 & 0.4491 & 0.4230 & 0.4086 & 0.3790 & 0.3497 & 0.4010 & 0.3819 & 0.3465 & \textbf{0.5902} & 0.4957 & 0.5484 & 0.5778 \\

    11    & 0.4162 & 0.4792 & \textbf{0.7862} & 0.7596 & 0.5085 & 0.4667 & 0.4266 & 0.4422 & 0.5359 & 0.4143 & 0.4699 & 0.5456 & 0.5781 & 0.6048 & 0.5948 \\

    12    & 0.0132 & 0.0046 & 0.0093 & 0.0077 & 0.0070 & 0.0023 & 0.0170 & 0.0000 & 0.0302 & 0.0152 & 0.0697 & 0.0511 & 0.0477 & \textbf{0.0927} & 0.0579 \\

    13    & 0.4525 & 0.5556 & 0.4238 & 0.4090 & 0.5442 & 0.5164 & 0.5707 & 0.5603 & 0.5324 & 0.4523 & 0.4789 & 0.5148 & 0.5619 & 0.4246 & \textbf{0.5811} \\

    14    & 0.3019 & 0.2629 & 0.5460 & 0.4665 & 0.3651 & 0.4152 & 0.2294 & 0.2073 & 0.3212 & 0.3789 & 0.3309 & 0.5289 & \textbf{0.6763} & 0.3375 & 0.5705 \\

    15    & 0.6303 & 0.4721 & 0.4457 & 0.4887 & 0.4602 & 0.5549 & 0.4180 & 0.5234 & 0.5944 & 0.5197 & 0.5289 & 0.6277 & 0.6476 & 0.6447 & \textbf{0.6591} \\

    16    & 0.6412 & 0.7611 & 0.6220 & 0.7648 & 0.7559 & 0.5888 & 0.7374 & 0.6403 & 0.6688 & 0.7457 & 0.7086 & 0.8498 & 0.7594 & 0.7173 & \textbf{0.8572} \\

    17    & \textbf{1.0000} & \textbf{1.0000} & \textbf{1.0000} & \textbf{1.0000} & \textbf{1.0000} & \textbf{1.0000} & 0.9545 & \textbf{1.0000} & \textbf{1.0000} & \textbf{1.0000} & \textbf{1.0000} & 0.9545 & 0.8909 & \textbf{1.0000} & \textbf{1.0000} \\

    18    & 0.4983 & 0.6885 & 0.6576 & 0.5197 & 0.7140 & 0.6581 & 0.6625 & 0.5686 & 0.7366 & 0.5346 & 0.6080 & 0.6365 & 0.5981 & \textbf{0.7692} & 0.7020 \\

    19    & 0.5265 & 0.6363 & 0.9060 & 0.8777 & 0.7323 & 0.6989 & 0.6100 & 0.6266 & 0.6771 & 0.6569 & 0.6545 & 0.9102 & 0.8960 & 0.8006 & \textbf{0.9476} \\

    20    & 0.4444 & 0.8904 & \textbf{0.9706} & 0.5253 & 0.8479 & 0.9189 & 0.6797 & 0.5955 & 0.5393 & 0.5388 & 0.4801 & 0.6246 & 0.5566 & 0.4879 & 0.6519 \\
    \midrule
    AA    & 0.5195 & 0.6039 & 0.5707 & 0.5696 & 0.6060 & 0.5952 & 0.5676 & 0.5647 & 0.5863 & 0.5777 & 0.5458 & 0.6400 & 0.6268 & 0.6090 & \textbf{0.6667} \\

    OA    & 0.4634 & 0.5101 & 0.5750 & 0.5899 & 0.5552 & 0.5027 & 0.4825 & 0.5492 & 0.5944 & 0.5938 & 0.4851 & 0.7278 & 0.6969 & 0.5116 & \textbf{0.7728} \\

    $\kappa$ & 0.3732 & 0.4317 & 0.4833 & 0.4974 & 0.4714 & 0.4231 & 0.4018 & 0.4560 & 0.5037 & 0.4948 & 0.3936 & 0.6474 & 0.6124 & 0.4372 & \textbf{0.7011} \\
    \bottomrule
    \end{tabular}%
    }
  \label{tab:HU2017}%
\end{table*}%

\begin{figure*} [tbp]\begin{center}
\begin{tabular}{cccccc}
\includegraphics[width=0.13\linewidth]{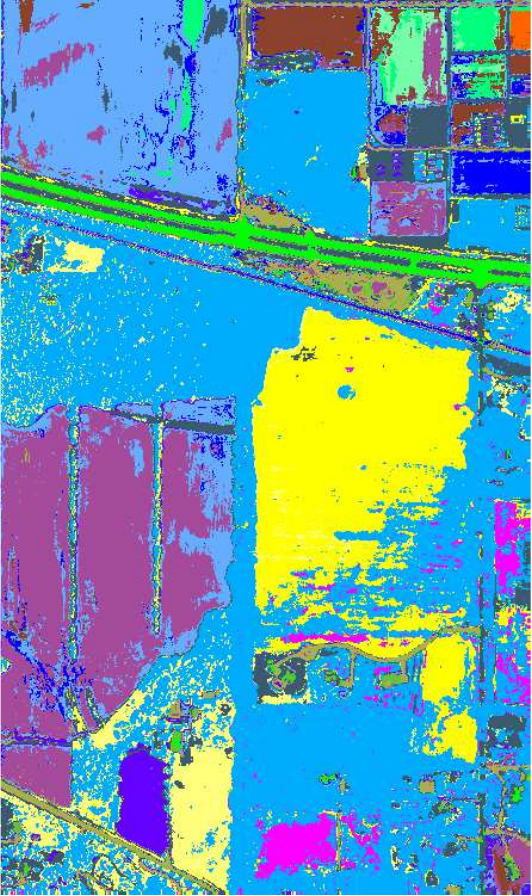}&\includegraphics[width=0.13\linewidth]{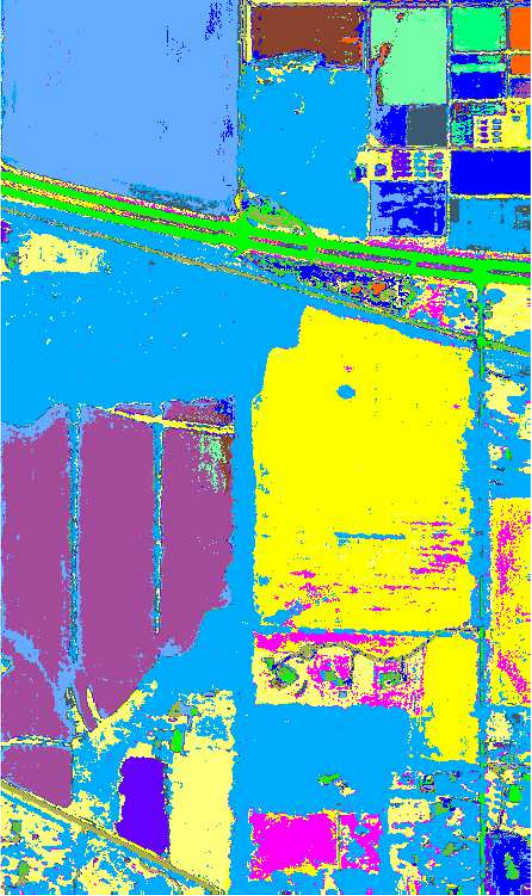}&\includegraphics[width=0.13\linewidth]{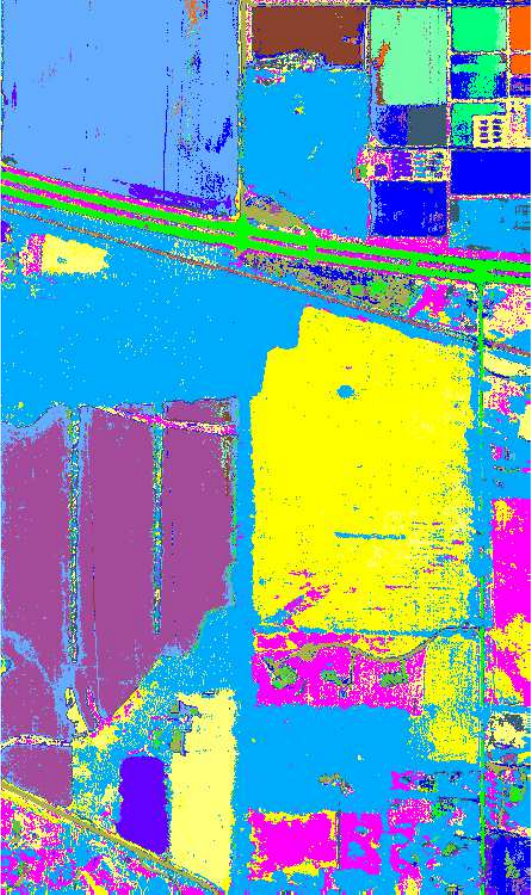}&\includegraphics[width=0.13\linewidth]{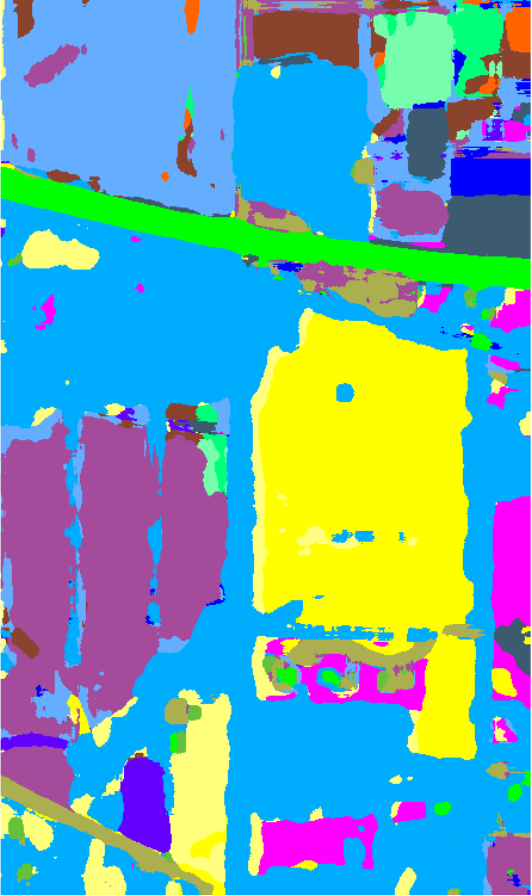} &\includegraphics[width=0.13\linewidth]{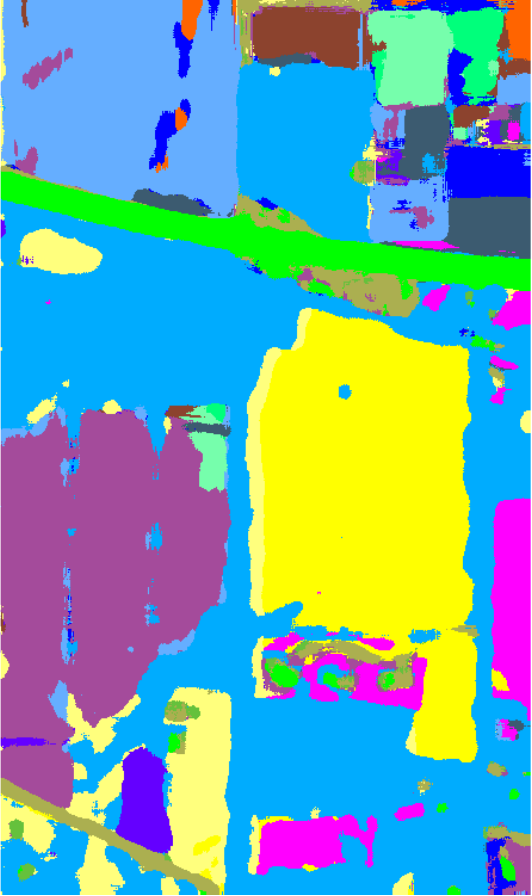}&\includegraphics[width=0.13\linewidth]{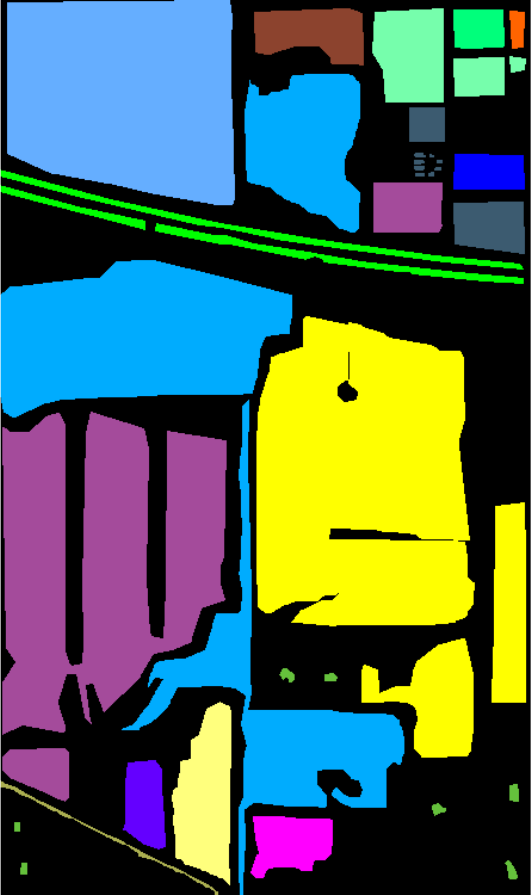}\\(a) HSI & (b) LPP & (c) JPlay & (d) CNN &(e) PCNN & (f) Ground  Reference\\
 \end{tabular} \end{center} \caption{Classification maps obtained  on the extracted features from Indian Pines 2010 data. From each category the method with highest OA is shown for the demonstration along with the one obtained from the spectral bands.}
 \label{fig:MapIP10}
\end{figure*}


\begin{figure*} [tbp]\begin{center}
\begin{tabular}{cc}
\includegraphics[width=0.48\linewidth]{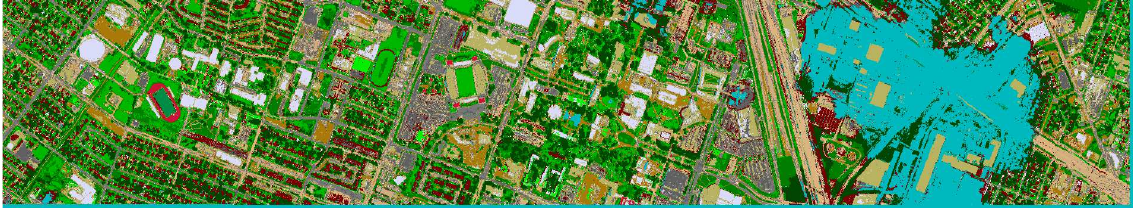}&\includegraphics[width=0.48\linewidth]{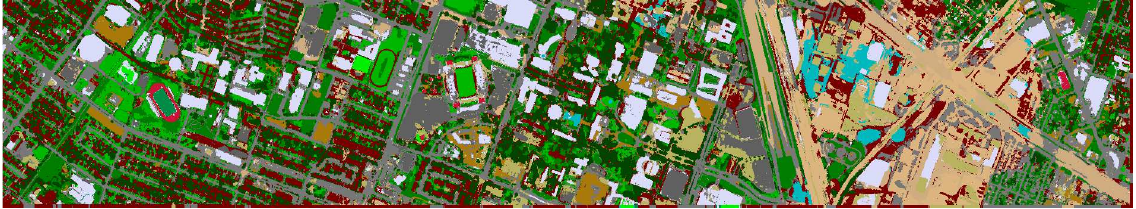}\\(a) HSI & (b) OTVCA \\\includegraphics[width=0.48\linewidth]{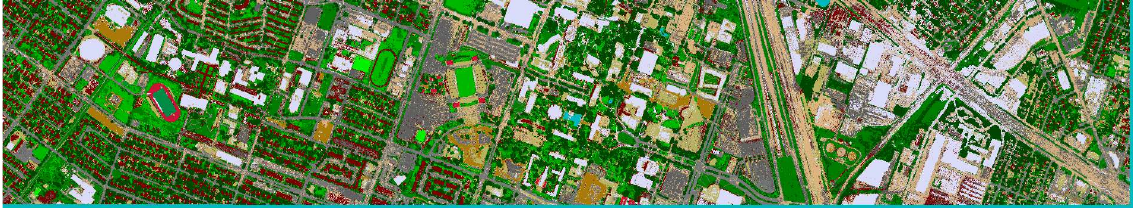}&\includegraphics[width=0.48\linewidth]{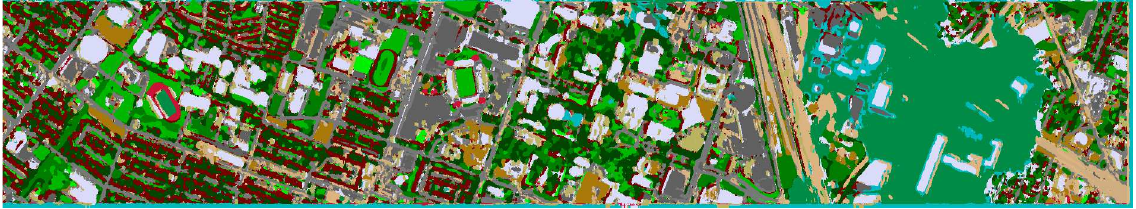} \\ (c) JPlay & (d) CNN\\\includegraphics[width=0.48\linewidth]{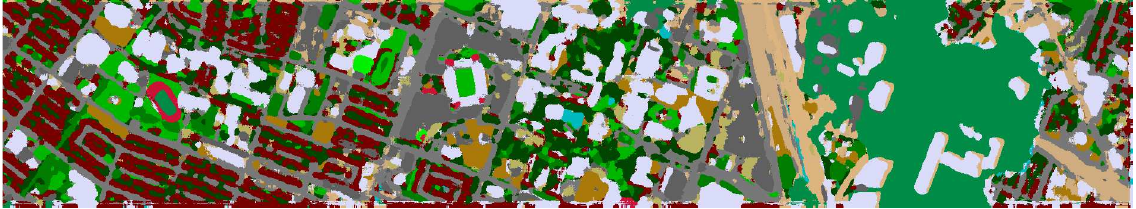}&\includegraphics[width=0.48\linewidth]{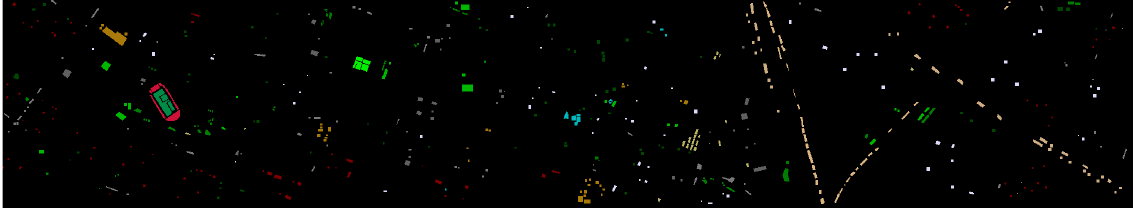} \\(e) PCNN & (f) Ground  Reference\\
 \end{tabular} \end{center} \caption{Classification maps obtained on the extracted features from Houston University 2013 data. From each category the method with highest OA is shown for the demonstration along with the one obtained from the spectral bands.}
 \label{fig:MapHU12}
\end{figure*}

\begin{figure*} [tbp]\begin{center}
\begin{tabular}{cc}
\includegraphics[width=0.48\linewidth]{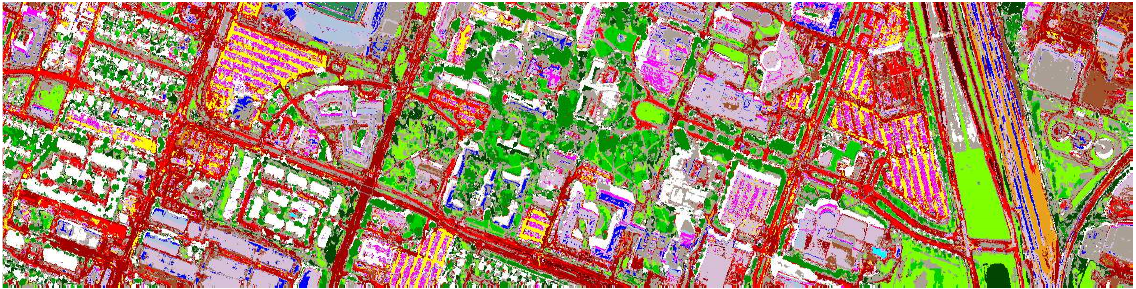}&\includegraphics[width=0.48\linewidth]{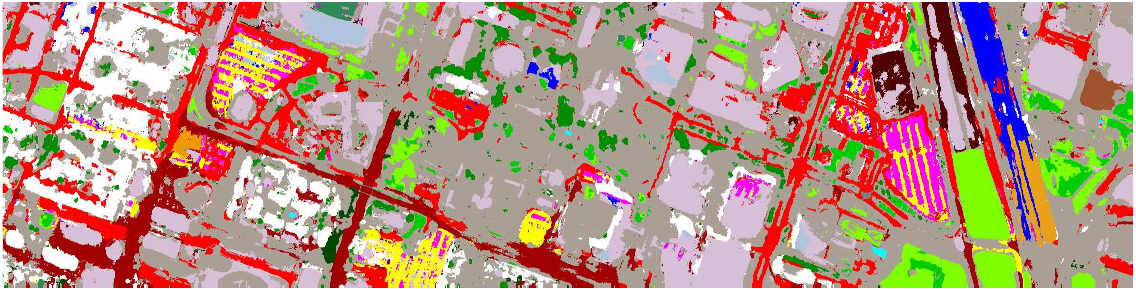}\\(a) HSI & (b) OTVCA \\\includegraphics[width=0.48\linewidth]{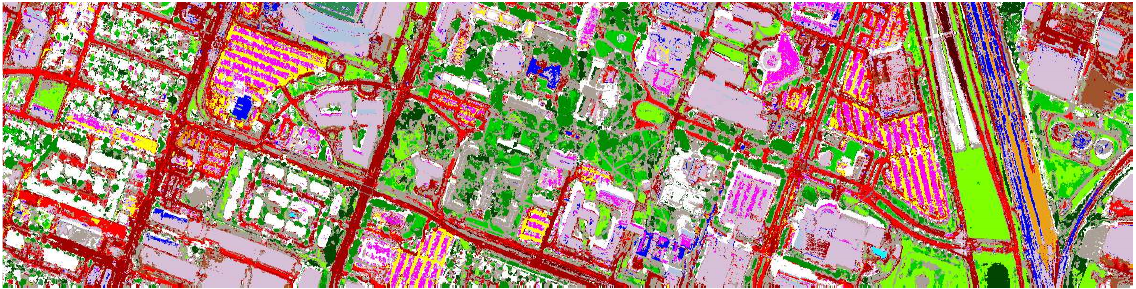}&\includegraphics[width=0.48\linewidth]{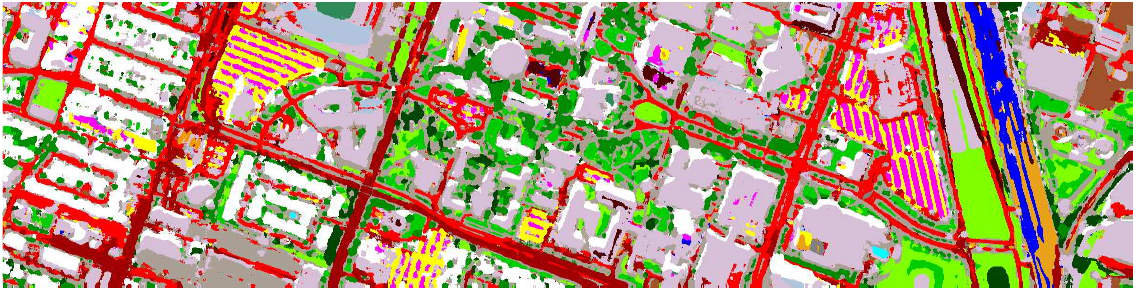} \\ (c) JPlay & (d) CNN\\\includegraphics[width=0.48\linewidth]{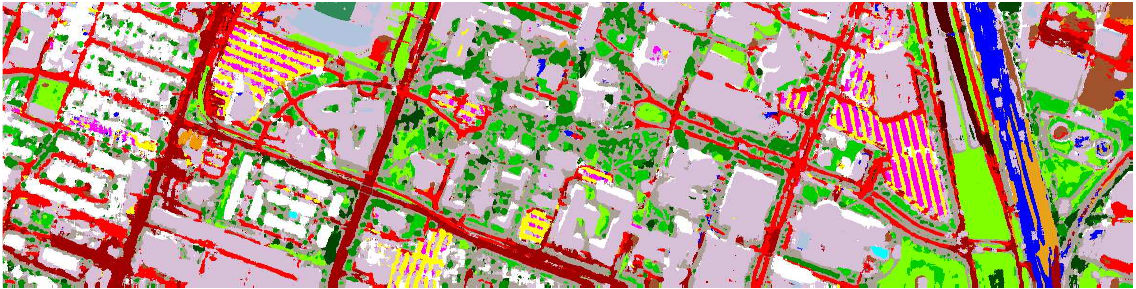}&\includegraphics[width=0.48\linewidth]{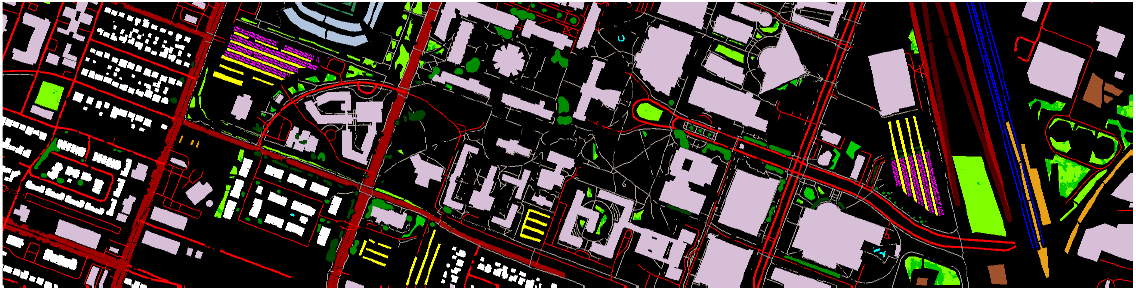} \\(e) PCNN & (f) Ground  Reference\\
 \end{tabular} \end{center} \caption{Classification maps obtained on the extracted features from Houston University 2018 data. From each category the method with highest OA is shown for the demonstration along with the one obtained from the spectral bands.}
 \label{fig:MapHU17}
\end{figure*}

\subsection{Performance of the FE techniques on three HSIs}
 We have applied the FE techniques on the three hyperspectral datasets, i.e., Indian Pinse 2010, Houston 2013, and Houston 2018, and the classification accuracies including class accuracies, AA, OA, and $\kappa$ coefficient are all shown in Tables \ref{tab:IP2010}, \ref{tab:HU2012}, and \ref{tab:HU2017}, respectively. The results are first discussed within the categories and then between different categories. We should note that the results and discussions are in terms of classification accuracies obtained from the classification of the HSIs.

\subsubsection{Unsupervised FE}
\begin{itemize}
    \item[$\triangleright$] PCA demonstrates the poorest performance compared with the other techniques, however, it considerably improves the classification accuracies compared with the results obtained by applying the RF on the spectral bands. One of the main disadvantages of the PCA is that, it does not take into account the noise and, therefore, the extracted features with having lower variance are often degraded by different types of noise existing in the hyperspectral image \cite{BR_Rev1}. Additionally, PCA only takes into account the spectral correlation and it entirely neglects the spatial (neighboring) information. 
     \item[$\triangleright$] LPP considerably outperforms the other UFE techniques for the Indian Pines dataset. However, in the case of Houston datasets, it provides very poor results. LPP incorporates the spatial information using the manifold learning process and by constructing the neighboring graph \cite{LPP}.
    \item[$\triangleright$] OTVCA outperforms the other UFE technqiues for Houston datasets. In the case of Houston 2013, the improvements are very considerable. OTVCA is robust to noise due to the signal model which takes into account the noise and model's errors. Additionally, OTVCA exploits the spatial correlation by incorporating the TV penalty and, therefore, the extracted features are piece-wise smooth and have high SNR \cite{OTVCA}. Overall, it can be observed that OTVCA, which is a candidate from the low-rank reconstruction techniques, generally provides better classification accuracies than other the UFE techniques.

\end{itemize}
\subsubsection{Supervised FE}
\begin{enumerate}
      \item[$\triangleright$] \textit{LDA versus Spectral Classifier (RF):} With the embedding of supervised information, LDA obviously performs better than the situation where RF is directly applied to the spectral signatures in terms of the overall performance and individual accuracies for most materials. This indicates the effectiveness of SFE to a great extent.
      \item[$\triangleright$] \textit{LDA versus CGDA:} Although the classification performance of CGDA is inferior to that of LDA from the whole perspective, yet the advantage of CGDA mainly lies in its automation in computing the similarity (or connectivity) between samples. This could lead to a relatively stable FE, particularly in large-scale and more complex hyperspectral scenes. Due to the data-driven graph embedding, CGDA yields a lower running speed than LDA in the process of model training.
      \item[$\triangleright$] \textit{LDA versus LSDR:} Intuitively, LSDR provides competitive classification performance with LDA. However, LSDR is time consuming due to the distribution matching between input samples and labels. The requirement to estimate the statistical distribution also limits the LSDR's stability, especially when the training set is available on a small scale (e.g., for the Indian  Pines 2010 and Houston 2013 datasets).
      \item[$\triangleright$] \textit{LDA versus JPlay:} Unlike the conventional regression techniques, JPlay is capable of extracting semantically meaningful and robust features, due to the multi-layered structure and the self-reconstruction constraint (\ref{SFE_eq18}). Quantitatively speaking, JPlay outperforms the other SFE methods. The CV provides a feasible solution to automatically determine the parameter combination in JPlay. Despite the ADMM solver designed for speeding up the optimization process such a multi-layered parameter update inevitably suffers from high computational cost.
\end{enumerate}

\subsubsection{Deep FE}
\begin{enumerate}

\item[$\triangleright$] \textit{Spectral versus Spectral-Spatial models:}
Most of the spectral-spatial models (i.e., CNN, PCNN, CAE, and CRNN) achieve superior performance than spectral models (i.e., SAE and RNN) in terms of AA, OA, and Kappa due to the joint use of spectral and spatial information. This indicates that, besides the rich spectral information, spatial information is also important for hyperspectral image classification.
\item[$\triangleright$] \textit{PCNN and CNN versus CAE and CRNN:}
Similar to SAE, CAE focuses on image reconstruction rather than classification. In contrast, PCNN and CNN are exclusively designed for the classification task, so they are able to learn more discriminative features than CAE, leading to better classification performance especially on the Houston 2018 dataset. Although CRNN also focuses on the classification task, it has a higher number of parameters to train. Using the same number of training samples and epochs, PCNN and CNN can achieve better results than CRNN in terms of AA, OA, and Kappa.
\item[$\triangleright$] \textit{PCNN versus CNN:}
PCNN outperforms CNN in terms of the classification accuracies for all three datasets. We should note that the improvements are substantial in the case of the Houston 2013 and 2018 datasets. Due to the use of PCA, most of the redundant spectral information is reduced. Therefore, the number of trainable parameters in PCNN is smaller than that of CNN, making it easier to learn under the same condition.
\end{enumerate}

\subsubsection{Shallow UFE Versus Shallow SFE}
For all three datasets used in the experiments, the UFE techniques provide better classification accuracies than the SFE techniques. Unlike the SFE, the UFE tends to pay more attentions on spatial-spectral information extraction, owing to fully considering all samples of HSI as the model input. Conversely, the performance of SFE is, to a great extent, limited by the ability to largely gather HSI ground sampling. A direct evident is given in Tables \ref{tab:IP2010}, \ref{tab:HU2012}, and \ref{tab:HU2017}. For the Indian Pines 2010 and Houston 2018 datasets where more training samples are available, SFE-based methods hold the competitive results with UFE-based ones, while for the Houston 2013 dataset, the classification performance of SFE is relatively inferior to that of UFE, due to the small-scale training set. Considering the low number of ground samples often available in HSI applications the experimental results confirm the advantage of UFE over SFE for HSI feature extraction.

\subsubsection{Shallow FE versus Deep FE}
At the first glance, the shallow FE approaches slightly outperform the deep FE techniques for the two datasets, i.e., Indian Pines 2010 and Houston 2013. However, a deep comparison reveals that some deep FE techniques, such as CNN-based FE, provide consistency and good performance over all three datasets. Additionally, when the dimension-reduced step (e.g., using PCA) is applied prior to the CNN technique, the resulting PCNN yields by far the second highest accuracies in the case of Indian Pines 2010 and Houston 2013 datasets (only moderately lower than LPP or OTVCA, respectively), and simultaneously obtains the best performance on the Houston 2018 dataset. It is worth mentioning that CNN-based FE methods obtain at least 10\% increase over the shallow ones in the case of Houston 2018. This could be due to the high nonlinear behaviour of this dataset which contains 20 land cover classes.
The main factors for CNN-based FE methods to obtain around 20\% of improvement over those shallow FE methods on the Houston 2018 dataset are the availability of the sufficient training samples and modeling the spatial information of HSI well.

\subsubsection{Comparisons of the Land Cover Maps}

Figs. \ref{fig:MapIP10}, \ref{fig:MapHU12}, and \ref{fig:MapHU17} compare the classification maps for Indian Pines, Houston University 2013, and Houston University 2018, respectively. The figures compare the maps obtained from methods which provide the highest OA from each category (i.e., Shallow UFE, shallow SFE, and deep FE) along with the map obtained from the spectral classifier (HSI). Additionally, we depicted the maps obtained by CNN for all three datasets since it provides the highest OA among the deep FE techniques which do not exploit a reduction step.

Overall, the classification maps of either unsupervised or supervised FE-based approaches (e.g., LPP, JPlay, CNN, PCNN) are smoother compared to the results only using HSI that tends to generate sparse mislabeled pixels. More specifically, the classification maps generated by spectral-spatial FE-based methods, e.g., OTVCA, CNN, and PCNN, are usually a bit over-smoothed, leading to the creation of fake structures, especially for the Indian  Pines 2010 and Houston 2018 datasets. In the case of OTVCA, the over smoothing can be avoided by decreasing the tuning parameter.
In contrast, JPlay obtains relatively desirable classification maps, despite the lack of spatial information modeling. It is worth mentioning that the JPlay algorithm can maintain the structural information  for Houston 2013 in the shadow covered region where pixels at some  bands are considerably attenuated. This is due to the elimination of the spectral variability using the self-reconstruction regularization (the third term in (\ref{SFE_eq18})) and the multi-layered linearized regression technique.

\subsection{Performance with respect to the Number of Training Samples}

\begin{figure} [tbp]\begin{center}
\begin{tabular}{c}
\includegraphics[width=0.8\linewidth]{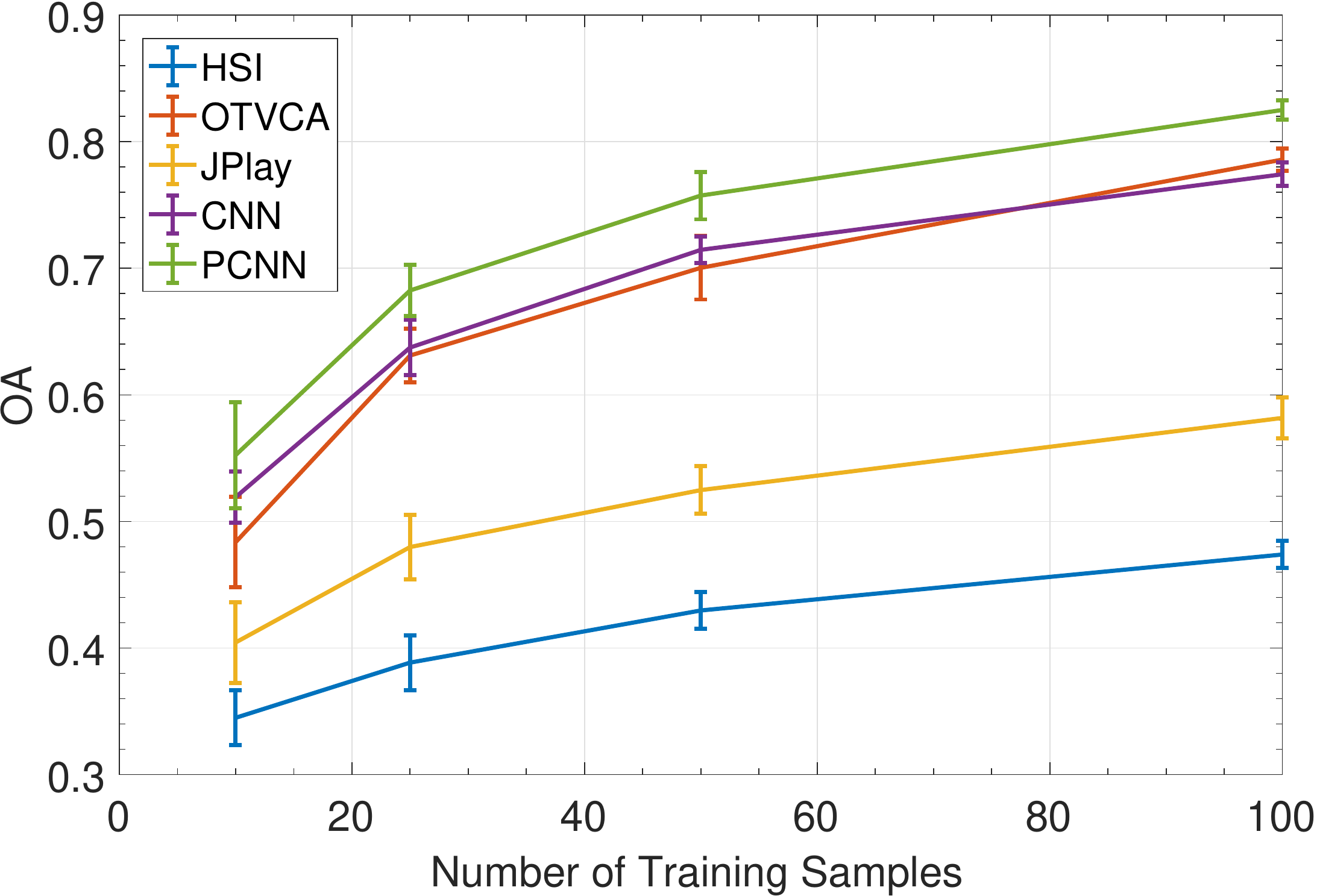}
 \end{tabular} \end{center} \caption{The classification accuracies w.r.t. the number of the training samples
 on Houston University 2018 data. The results shown are means over 10 experiments and the standard deviations are shown by the error bars.}
 \label{fig:OA_No}
\end{figure}

In this subsection, we investigate the performance of the FE techniques in terms of classification accuracies with respect to the number of training samples. As we have already stated, this analysis is of great interest due to two main reasons; first, ground sample acquisition and measurements are often cumbersome and could be impossible in cases for which the target area is not reachable. Additionally, the limited number of samples not only affects the performance of the supervised classifiers but also the supervised features extraction techniques since they are highly reliable on the number of training samples. Therefore, in this experiment, we perform an analysis on the Houston University dataset 2018 by comparing the performances of the feature extraction techniques by selecting 10, 25, 50, and 100 training samples randomly.
Fig \ref{fig:OA_No} compares the OAs obtained by applying RF on the spectral bands (labeled by HSI), the features extracted by OTVCA, and JPlay along with the OAs obtained by CNN, and PCNN.
The results are mean values over 10 experiments by selecting the samples randomly (the standard deviations are shown by the error bars). The outcomes of the experiment can be summarized as follows:

\begin{itemize}
    \item[$\triangleright$] The SFE technique (i.e., JPlay) improves the OAs compared to the spectral classifier. However, it provides much lower OA compared with the UFE and deep FE for all cases. Two aspects might explain this point. One is that JPlay fails to model spatial and contextual information, and another is that although JPlay attempts to enhance the reorientation ability of the features via multi-layered linear mapping, yet it is still incomparable to the nonlinear deep FE-based techniques, particularly when the number of samples are increased.
    \item[$\triangleright$] In this experiment, the UFE technique (i.e. OTVCA) and the deep FE one, CNN, are performed similarly in terms of classification accuracies. Compared with the results given in Table \ref{tab:HU2017}, it can be observed that the random selection of the training samples over the entire class regions from the ground reference considerably improves the performance of RF applied on the features extracted by OTVCA. This is often due to the lack of a parameter selection technique to select the optimum parameter for the OTVCA algorithm which could lead to over-smoothing on the features. 
    \item[$\triangleright$] The deep learning technique (i.e., PCNN) after using the reduction (i.e., PCA) provides very high OA for all the cases. Comparing the results with CNN (i.e., without using the PCA reduction) confirms the advantage of using the reduction stage prior to deep learning techniques.
\end{itemize}



\section{Conclusion and Summery}
\label{sec:con}
In the past decade, hyperspectral image FE has considerably evolved leading to three main research lines (i.e., shallow unsupervised, shallow supervised, and deep feature extraction approaches) that include the majority of feature extraction techniques, which were studied in this paper. The paper has systematically provided a technical overview of the state-of-the-art techniques proposed in the literature by categorizing the aforementioned three focuses into subcategories. In order to make this
research paper easy-to-follow for researchers at different levels (i.e., students, researchers, and senior researchers), we aimed at showing the evolution of each category over the decades rather than including many techniques with an exhaustive reference list. The experimental section was designed in a way to compare the performances of the techniques in two ways: (1) between all the categories (i.e., shallow unsupervised, shallow supervised, and deep feature extraction approaches) and (2) within each category by analyzing the corresponding subcategories. In this manner, a variety of subcategories is investigated detailing the evolution of shallow unsupervised approaches (i.e., conventional data projection schemes, band clustering/splitting techniques, low-rank reconstruction techniques, and manifold learning techniques), shallow supervised approaches (i.e., class separation discriminant analysis,  graph embedding discriminant analysis,  regression-based representation learning, and joint \& progressive learning strategy), and deep approaches (i.e., AE, CNN, RNN, and integrative approaches). Three recent hyperspectral datasets have been studied in this paper and the results are evaluated in terms of classification accuracies and the quality of the classification maps. 
The experiments carried out in this study showed that in terms of classification accuracies; 1) deep learning
FE techniques (i.e., CNN and PCNN) can outperform the shallow ones in particular when a sufficient amount of training data is available 2) applying a dimensionality reduction step (such as PCA) prior to the deep learning technqiues considerably improves their performances 3) shallow UFE techniques not only outperform the SFE ones but also are very competitive compared with deep FE ones. However, we should mention that the conclusions are limited by the experiments carried out on the three HSI datasets. In addition, this paper provides an impressive amount of codes and libraries mostly written in Python and Matlab to ease out the task of researchers in this very vibrant field of research.

%

\section*{Acknowledgment}
We would like to thank Prof. Melba Crawford for providing the Indian Pines 2010 Data and the National Center
for Airborne Laser Mapping, the Hyperspectral Image
Analysis Laboratory at the University of Houston, and
the IEEE GRSS Image Analysis and Data Fusion Technical
Committee. This work is partially supported by an Alexander von Humboldt research grant. We also would like to thank the AXA Research Fund for supporting the work of
Prof. Jocelyn Chanussot. The corresponding author of this paper is Dr. Danfeng Hong.

\ifCLASSOPTIONcaptionsoff
  \newpage
\fi

\bibliographystyle{IEEEtran}
\bibliography{refs}

\end{document}